\documentstyle[rotate,fleqn,fullname]{article}

\title{Principles and Implementation of\\Deductive Parsing}

\author{
  Stuart M. Shieber \\
  Division of Applied Sciences\\
  Harvard University, Cambridge, MA 02138
\and
  Yves Schabes\\
  Mitsubishi Electric Research Laboratories\\
  Cambridge, MA 02139
\and
  Fernando C. N. Pereira\\
  AT\&T Bell Laboratories\\
  Murray Hill, NJ 07974
}

\date{\today}

\newcommand{\codeincluded}[2]{#1}	% if code is included
\input{psfig-scale}

\def\term{\begingroup\tt\catcode`\_=12 \obeyspaces\doterm}
\def\doterm#1{#1\endgroup}

\def\beginprolog{\par\vspace{-.5ex}\begingroup\leftskip=\leftmargini\tt
		 \setlength{\parindent}{0in}
		 \setlength{\parskip}{0in}
	         \def\par{\leavevmode\endgraf}	% from TEX manual p. 381
		 \catcode`\^=12	% make specials unspecial
		 \catcode`\&=12 %
		 \catcode`\_=12 %
                 \catcode`\|=12 %
		 \obeylines%
		 \catcode`\ =\active%
		 \catcode`\%=12\catcode`\`=\active}
{\catcode`\ =\active\global\let =\enskip}
{\catcode`\`=\active\gdef`{\relax\lq}}

\def\endprolog{\endgroup\par\vspace{2.2ex}}

\def\prologlisting#1{\vspace{2.3ex}\beginprolog\input#1\endprolog\vspace{-.3ex}}

\newcommand{\oneovermodb}[3]{\begin{array}[b]{@{}c@{}}
                                   #2 \\[-1.8ex]
                                   \hbox to #1{\hrulefill} \\[-.8ex]
                                   #3 \end{array}}

\newcommand{\oneover}[2]{\begin{array}{@{}c@{}} #1 \\
\hline #2  \end{array}}

\newcommand{\oneovermod}[3]{\begin{array}{@{}c@{}}
                                   #2 \\[-1.8ex]
                                   \hbox to #1{\hrulefill} \\[-.8ex]
                                   #3 \end{array}}

\newcommand{\note}[2]{}
\newcommand{\eqpunc}[1]{{\makebox[0pt][l]{\qquad\rm{#1}}}}
\newcommand{\seq}[1]{\langle #1 \rangle}

\newcommand{\ra}{\rightarrow}
\newcommand{\Ra}{\Rightarrow}
\newcommand{\derivestar}{\stackrel{*}{\Rightarrow}}
\newcommand{\den}[1]{[\![ #1 ]\!]}

\newcommand{\userule}[2]{\mbox{{\sc #1} from #2}}
\newcommand{\useaxiom}{\mbox{{\sc axiom}}}

\newcommand{\predict}[1]{\userule{predict}{#1}}
\newcommand{\scan}[1]{\userule{scan}{#1}}
\newcommand{\complete}[2]{\userule{complete}{#1 and #2}}
\newcommand{\shift}[1]{\userule{shift}{#1}}
\newcommand{\reduce}[1]{\userule{reduce}{#1}}

\newcommand{\bk}{\mbox{$\backslash$}}
\newcommand{\fw}{\mbox{/}}

\newcommand{\NP}{\mbox{\it NP}}

\newcommand{\ccgrule}[2]{$#2$ &$\ra$& $#1$}

\newcommand{\posa}{^\bullet}
\newcommand{\posb}{_\bullet}
\newcommand{\address}{p}

\newcommand{\anode}[2]{#1{@}#2}
\newcommand{\subtree}[2]{#1/#2}
\newcommand{\nodelab}[1]{\mbox{\it Label}(#1)}
\newcommand{\treefoot}[1]{\mbox{\it Foot}(#1)}
\newcommand{\adjoinable}[1]{\mbox{\it Adj}(#1)}

\begin{document}

\begin{titlepage}
\maketitle
\thispagestyle{empty}

\begin{abstract}

We present a system for generating parsers based directly on the
metaphor of parsing as deduction.  Parsing algorithms can be
represented directly as deduction systems, and a single deduction
engine can interpret such deduction systems so as to implement the
corresponding parser.  The method generalizes easily to parsers for
augmented phrase structure formalisms, such as definite-clause
grammars and other logic grammar formalisms, and has been used for
rapid prototyping of parsing algorithms for a variety of formalisms
including variants of tree-adjoining grammars, categorial grammars,
and lexicalized context-free grammars.

\end{abstract}

\vfill
{\noindent\small
This paper is available from the Center for Research in Computing
Technology,  Division of Applied Sciences, Harvard University
as Technical Report TR-11-94, and through the Computation and Language
e-print archive as cmp-lg/9404008.}

\end{titlepage}

\section{Introduction}
\label{sec:intro}

Parsing can be viewed as a deductive process that seeks to prove
claims about the grammatical status of a string from assumptions
describing the grammatical properties of the string's elements and the
linear order between them.  Lambek's syntactic calculi
\cite{Lambek:sentstruct} comprise an early formalization of this
idea, which more recently was explored in relation to grammar
formalisms based on definite clauses
\cite{Colmerauer:mgs,Pereira+Warren:DCGs,Pereira+Warren:ED} and on
feature logics
\cite{Shieber:inference,Rounds+Manaster-Ramer:fg,Carpenter:logic}.

The view of parsing as deduction adds two main new sources of insights
and techniques to the study of grammar formalisms and parsing:
\begin{enumerate}
\item Existing logics can be used as a basis for new grammar
formalisms with desirable representational or computational
properties.
\item The modular separation of parsing into a logic of
grammaticality claims and a proof search procedure allows the
investigation of a wide range of parsing algorithms for existing
grammar formalisms by selecting specific classes of grammaticality
claims and specific search procedures.
\end{enumerate}
While most of the work on deductive parsing has been concerned with
(1), we will in this paper investigate (2), more specifically how to
synthesize parsing algorithms by combining specific logics of
grammaticality claims with a fixed search procedure.  In this way,
deduction can provide a metaphor for parsing that encompasses a wide
range of parsing algorithms for an assortment of grammatical
formalisms.  We flesh out this metaphor by presenting a series of
parsing algorithms literally as inference rules, and by providing a
uniform deduction engine, parameterized by such rules, that can be
used to parse according to any of the associated algorithms.  The
inference rules for each logic will be represented as unit clauses and
the fixed deduction procedure, which we provide a Prolog
implementation of, will be a version of the usual bottom-up
consequence closure operator for definite clauses. As we will show,
this method directly yields dynamic-programming versions of standard
top-down, bottom-up, and mixed-direction (Earley) parsing procedures.
In this, our method has similarities with the use of pure bottom-up
deduction to encode dynamic-programming versions of definite-clause
proof procedures in deductive databases
\cite{Bancilhon+Ramakrishnan:amateur,Naughton+Ramakrishnan:bottom-up}.

The program that we develop is especially useful for rapid prototyping
of and experimentation with new parsing algorithms, and was in fact
developed for that purpose.  We have used it, for instance, in the
development of algorithms for parsing with tree-adjoining grammars,
categorial grammars, and lexicalized context-free grammars.

Many of the ideas that we present are not new.  Some have been
presented before; others form part of the folk wisdom of the logic
programming community.  However, the present work is to our knowledge
the first to make the ideas available explicitly in a single notation
and with a clean implementation.  In addition, certain observations
regarding efficient implementation may be novel to this work.

The paper is organized as follows: After reviewing some basic logical
and grammatical notions and applying them to a simple example
(Section~\ref{sec:pasd}), we describe how the structure of a variety
of parsing algorithms for context-free grammars can be expressed as
inference rules in specialized logics (Section~\ref{sec:cf}).  Then,
we extend the method for stating and implementing parsing algorithms
for formalisms other than context-free grammars
(Section~\ref{sec:others}).  Finally, we discuss how deduction should
proceed for such logics, developing an agenda-based deduction
procedure implemented in Prolog that manifests the presented ideas
(Section~\ref{sec:control}).

\section{Basic Notions}

\label{sec:pasd}

As introduced in Section~\ref{sec:intro}, we see parsing as a
deductive process in which rules of inference are used to derive
statements about the grammatical status of strings from other such
statements. Statements are represented by formulas in a suitable
formal language.  The general form of a rule of inference is
\[ \oneover{A_1 \quad \cdots \quad A_k}
           {B} \quad \mbox{$\langle${side conditions on
$A_1,\ldots,A_k,B$}$\rangle$}
		 \eqpunc{.}
\]
The {\em antecedents} $A_1,\ldots,A_k$ and the {\em consequent} $B$ of
the inference rule are formula schemata, that is, they may contain
syntactic metavariables to be instantiated by appropriate terms when
the rule is used.  A grammatical deduction system is defined by a set
of rules of inference and a set of {\em axioms} given by appropriate
formula schemata.

Given a grammatical deduction system, a {\em derivation} of a formula
$B$ from assumptions $A_1,\ldots,A_m$ is, as usual, a sequence of
formulas $S_1,\ldots,S_n$ such that $B=S_n$, and each $S_i$ is either
an axiom (one of the $A_j$) or there is a rule of inference $R$ and
formulas $S_{i_1},\ldots,S_{i_k}$ with $i_1,\ldots,i_k<i$ such that
for appropriate substitutions of terms for the metavariables in $R$,
$S_{i_1},\ldots,S_{i_k}$ match the antecedents of the rule, $S_i$
matches the consequent, and the rule's side conditions are satisfied.
We write $A_1,\ldots,A_m\vdash B$ and say that $B$ is a {\em
consequence} of $A_1,\ldots,A_m$ if such a derivation exists. If $B$
is a consequence of the empty set of assumptions, it is said to be
{\em derivable}, in symbols $\vdash B$.

In our applications of this model, rules and axiom schemata may refer
in their side conditions to the rules of a particular grammar, and
formulas may refer to string positions in the fixed string to be
parsed $w = w_1 \cdots w_n$. With respect to the given string, {\em
goal formulas} state that the string is grammatical according to the
given grammar. Then parsing the string corresponds to finding a
derivation witnessing a goal formula.

We will use standard notation for metavariables ranging over the
objects under discussion: $n$ for the length of the object language
string to be parsed; $A, B, C \ldots$ for arbitrary formulas or
symbols such as grammar nonterminals; $a, b, c, \ldots$ for arbitrary
terminal symbols; $i,j,k,\ldots$ for indices into various strings,
especially the string $w$; $\alpha, \beta, \gamma, \ldots$ for strings
or terminal and nonterminal symbols.  We will often use such notations
leaving the type of the object implicit in the notation chosen for it.
Substrings will be notated elliptically as, e.g., $w_i \cdots w_j$ for
the $i$-th through $j$-th elements of $w$, inclusive.  As is usual, we
take $w_i \cdots w_j$ to be the empty string if $i > j$.

\subsection{A First Example: CYK Parsing}
\label{sec:cyk}

As a simple example, the basic mechanism of the Cocke-Younger-Kasami
(CYK) context-free parsing algorithm \cite{ka65,y67} for a
context-free grammar in Chomsky normal form can be easily represented
as a grammatical deduction system.

We assume that we are given a string $w=w_1\cdots w_n$ to be parsed
and a context-free grammar $G = \seq{N,\Sigma, P, S}$ , where $N$ is
the set of nonterminals including the start symbol $S$, $\Sigma$ is
the set of terminal symbols, ($V = N \cup \Sigma$ is the vocabulary of
the grammar,) and $P$ is the set of productions, each of the form $A
\ra \alpha$ for $A \in N$ and $\alpha \in V^*$.  We will use the
symbol $\Ra$ for immediate derivation and $\derivestar$ for its
reflexive, transitive closure, the derivation relation.  In the case
of a Chomsky-normal-form grammar, all productions are of the form $A
\ra B\; C$ or $A \ra a$.

The {\em items} of the logic (as we will call parsing logic formulas
from now on) are of the form $[A, i, j]$, and state that the
nonterminal $A$ derives the substring between indices $i$ and $j$ in
the string, that is, $A \derivestar w_{i+1}
\cdots w_j$.  Sound axioms, then, are grounded in the lexical items
that occur in the string.  For each word $w_{i+1}$ in the string and
each rule $A \ra w_{i+1}$, it is clear that the item $[A,i,i+1]$ makes
a true claim, so that such items can be taken as axiomatic.  Then
whenever we know that $B \derivestar w_{i+1} \cdots w_j$ and $C
\derivestar w_{j+1} \cdots w_k$ --- as asserted by items of the form
$[B,i,j]$ and $[C,j,k]$ --- where $A \ra B\; C$ is a production in the
grammar, it is sound to conclude that $C \derivestar w_{i+1} \cdots
w_k$, and therefore, the item $[C,i,k]$ should be inferable.  This
argument can be codified in a rule of inference:
\[ \oneover{[B,i,j] \qquad [C,j,k]}
           {[A,i,k]}
		\quad \mbox{ $A \ra B\; C$}
\]
Using this rule of inference with the axioms, we can conclude that the
string is admitted by the grammar if an item of the form $[S,0,n]$ is
deducible, since such an item asserts that $S \derivestar w_1\cdots
w_n = w$.  We think of this item as the {\em goal item} to be proved.

In summary, the CYK deduction system (and all the deductive parsing
systems we will define) can be specified with four components: a class
of items; a set of axioms; a set of inference rules; and a subclass of
items, the goal items.  These are given in summary form in
Figure~\ref{fig:cyk-sys}.

\begin{figure}
\begin{center}
\begin{tabular}{lll}

{\bf Item form:}	&	 $[A,i,j]$	\\ \\

{\bf Axioms:}		&	 $[A,i,i+1] \quad
				\mbox{ $A \ra w_{i+1}$}$ \\ \\

{\bf Goals:}		&	 $[S,0,n]$	\\ \\

{\bf Inference rules:}	&
	\( \oneover{[B,i,j] \qquad [C,j,k]}
		   {[A,i,k]}
			\quad \mbox{ $A \ra B\ C$} \)

\end{tabular}
\end{center}
\caption{The CYK deductive parsing system.}
\label{fig:cyk-sys}
\end{figure}

This deduction system can be encoded straightforwardly by the
following logic program:
\beginprolog
nt(A, I1, I) :-
    word(I, W),
    (A ---> [W]),
    I1 is I - 1.
nt(A, I, K) :-
    nt(B, I, J),
    nt(C, J, K),
    (A ---> [B, C]).
\endprolog
\noindent where {\tt $A$ ---> [$X_1$,$\ldots$,$X_m$]} is the encoding
of a production $A\ra X_1\cdots X_n$ in the grammar and
{\tt word($i$,$w_i$)} holds for each input word $w_i$ in the string to be
parsed. A suitable bottom-up execution of this program, for example
using the {\em semi-na\"{\i}ve} bottom-up procedure
\cite{Naughton+Ramakrishnan:bottom-up} will behave similarly to the
CYK algorithm on the given grammar.

\subsection{Proofs of Correctness}

Rather than implement each deductive system like the CYK one as a
separate logic program, we will describe in Section \ref{sec:control}
a meta-interpreter for logic programs obtained from grammatical
deduction systems. The meta-interpreter is just a variant of the
semi-na\"{\i}ve procedure specialized to programs implementing
grammatical deduction systems. We will show in Section
\ref{sec:control} that our procedure generates only items derivable
from the axioms ({\em soundness}) and will enumerate all the derivable
items ({\em completeness}).  Therefore, to show that a particular
parsing algorithm is correctly simulated by our meta-interpreter, we
basically need to show that the corresponding grammatical deduction
system is also sound and complete with respect to the intended
interpretation of grammaticality items. By sound here we mean that
every derivable item represents a true grammatical statement under
the intended interpretation, and by complete we mean that the item
encoding every true grammatical statement is derivable. (We also need
to show that the grammatical deduction system is faithfully
represented by the corresponding logic program, but in general this
will be obvious by inspection.)

\section{Deductive Parsing of Context-Free Grammars}
\label{sec:cf}
We begin the presentation of parsing methods stated as deduction
systems with several standard methods for parsing context-free
grammars.  In what follows, we assume that we are given a string $w =
w_1 \cdots w_n$ to be parsed along with a context-free grammar $G =
\seq{N,\Sigma, P, S}$.

\subsection{Pure Top-Down Parsing (Recursive Descent)}

The first full parsing algorithm for arbitrary context-free grammars
that we present from this logical perspective is recursive-descent
parsing.  Given a context-free grammar $G = \seq{N, \Sigma, P, S}$,
and a string $w = w_1 \cdots w_n$ to be parsed, we will consider a
logic with items of the form $[ {} \bullet \beta, j ]$ where $0 \leq j
\leq n$.  Such an item asserts that the substring of the string $w$ up
to and including the $j$-th element, when followed by the string of
symbols $\beta$, forms a sentential form of the language, that is,
that $S \derivestar w_1 \cdots w_j \beta$.  Note that the dot in the
item is positioned just at the break point in the sentential form
between the portion that has been recognized (up through index $j$)
and the part that has not ($\beta$).

Taking the set of such items to be the [propositional] formulas of the
logic, and taking the informal statement concluding the previous
paragraph to provide a denotation for the sentences,\footnote{A more
formal statement of the semantics could be given, e.g., as
\[ \den{[ {} \bullet \beta, j ]} = \left\{\begin{array}{ll}
     \mbox{\it truth} & \mbox{if $S \derivestar w_1 \cdots w_j \beta$}\\
     \mbox{\it falsity} & \mbox{otherwise}\eqpunc{.}
   \end{array} \right. \]}
we can explore a proof theory for the logic.  We start with an axiom
\[[{} \bullet S, 0 ]\eqpunc{,}\]
which is sound because $S \derivestar S$ trivially.

Note that two items of the form $[ {} \bullet w_{j+1} \beta, j ]$ and
$[ {} \bullet \beta, j+1 ]$ make the same claim, namely that $S
\derivestar w_1 \cdots w_j w_{j+1} \beta$.  Thus, it is clearly sound
to conclude the latter from the former, yielding the inference rule:
\[ \oneover{[ {} \bullet w_{j+1} \beta, j ]}
           {[ {} \bullet \beta, j+1 ]}\eqpunc{,}
\]
which we will call the {\it scanning} rule.

A similar argument shows the soundness of the {\it prediction} rule:
\[ \oneover{[ {} \bullet B \beta, j ]}
           {[ {} \bullet \gamma \beta, j ]}
                \quad \mbox{ $B \ra \gamma$}\eqpunc{.} \]

Finally, the item $[ {} \bullet {}, n ]$ makes the claim that $S
\derivestar w_1 \cdots w_n$, that is, that the string $w$ is admitted
by the grammar.  Thus, if this goal item can be proved from the axiom
by the inference rules, then the string must be in the grammar.  Such
a proof process would constitute a sound recognition algorithm.  As it
turns out, the recognition algorithm that this logic of items
specifies is a pure top-down left-to-right regime, a recursive-descent
algorithm. The four components of the deduction system for top-down
parsing --- class of items, axioms, inference rules, and goal items
--- are summarized in Figure~\ref{fig:td-sys}.

\begin{figure}
\begin{center}
\begin{tabular}{ll}

{\bf Item form:}	&	 $[ {} \bullet \beta, j ]$	\\ \\

{\bf Axioms:}		&	 $[ {} \bullet S, 0 ]$		\\ \\

{\bf Goals:}		&	 $[ {} \bullet {}, n ]$		\\ \\

{\bf Inference rules:}\\

{\bf \quad Scanning} &

	\( \oneover{[ {} \bullet w_{j+1} \beta, j ]}
		   {[ {} \bullet \beta, j+1 ]} \)		\\ \\

{\bf \quad Prediction} &

	\( \oneover{[ {} \bullet B \beta, j ]}
		   {[ {} \bullet \gamma \beta, j ]}
			\quad \mbox{ $B \ra \gamma$} \)

\end{tabular}
\end{center}
\caption{The top-down recursive-descent deductive parsing system.}
\label{fig:td-sys}
\end{figure}

To illustrate the operation of these inference rules for context-free
parsing, we will use the toy grammar of Figure \ref{fig:toy}.
\begin{figure}
\begin{center}
\begin{tabular}{ll}
\parbox[t]{2in}{
\begin{eqnarray*}
S&\ra& \mbox{\it NP}\; \mbox{\it VP} \\
\mbox{\it NP}&\ra& \mbox{\it Det}\; N\; \mbox{\it OptRel} \\
\mbox{\it NP}&\ra& \mbox{\it PN} \\
\mbox{\it VP}&\ra& \mbox{\it TV}\; \mbox{\it NP} \\
\mbox{\it VP}&\ra& \mbox{\it IV} \\
\mbox{\it OptRel}&\ra& \mbox{\it RelPro}\; \mbox{\it VP} \\
\mbox{\it OptRel}& \ra &\epsilon
\end{eqnarray*}} &
\parbox[t]{2in}{\begin{eqnarray*}
\mbox{\it Det}&\ra& \mbox{a} \\
N &\ra&\mbox{program}\\
\mbox{\it PN}&\ra& \mbox{Terry} \\
\mbox{\it PN}&\ra& \mbox{Shrdlu} \\
\mbox{\it IV}&\ra& \mbox{halts} \\
\mbox{\it TV}&\ra& \mbox{writes} \\
\mbox{\it RelPro}&\ra& \mbox{that}
\end{eqnarray*}}
\end{tabular}
\end{center}
\caption{An example context-free grammar.}
\label{fig:toy}
\end{figure}
Given that grammar and the string
\begin{equation}
w_{1}w_{2}w_{3} =\mbox{a program halts} \label{eg-sent}
\end{equation}
we can construct the following derivation using the rules just given:
\[
\begin{tabular}{rll}
1 & $[ {} \bullet S, 0 ]$ & \useaxiom \\
2 & $[ {} \bullet \mbox{\it NP}\; \mbox{\it VP}, 0 ]$ & \predict{1} \\
3 & $[ {} \bullet  \mbox{\it Det}\; N\; \mbox{\it OptRel}\;\mbox{\it VP},0 ]$ &
\predict{2} \\
4 & $[ {} \bullet  \mbox{a}\; N\; \mbox{\it OptRel}\; \mbox{\it VP},0 ]$ &
\predict{3} \\
5 & $[ {} \bullet N\; \mbox{\it OptRel}\; \mbox{\it VP},1 ]$ & \scan{4} \\
6 & $[ {} \bullet \mbox{program}\; \mbox{\it OptRel}\; \mbox{\it VP},1 ]$ &
\predict{5} \\
7 & $[ {} \bullet \mbox{\it OptRel}\; \mbox{\it VP},2 ]$ & \scan{6} \\
8 & $[ {} \bullet \mbox{\it VP},2 ]$ & \predict{7} \\
9 & $[ {} \bullet \mbox{\it IV},2 ]$ & \predict{8} \\
10 & $[ {} \bullet \mbox{halts},2 ]$ & \predict{9} \\
11 & $[ {} \bullet {},3 ]$ & \scan{10}
\end{tabular}
\]
The last item is a goal item, showing that the given sentence is
accepted by the grammar of Figure~\ref{fig:toy}.

The above derivation, as all the others we will show, contains just
those items that are strictly necessary to derive a goal item from
the axiom. In general, a complete search procedure, such as the one we
describe in Section \ref{sec:control}, generates items that are either
dead-ends or redundant for a proof of grammaticality. Furthermore,
with an ambiguous grammar there will be several essentially different
proofs of grammaticality, each corresponding to a different analysis
of the input string.

\subsubsection{Proof of Completeness}

We have shown informally above that the inference rules for top-down
parsing are sound, but for any such system we also need the guarantee
of {\em completeness}: if a string is admitted by the grammar, then
for that string there is a derivation of a goal item from the initial
item.

In order to prove completeness, we prove the following lemma: If
$S\derivestar w_1 \cdots w_j \gamma$ is a leftmost derivation (where
$\gamma \in V^*$), then the item $[{}\bullet\gamma,j]$ is generated.
We must prove all possible instances of this lemma.  Any specific
instance can be characterized by specifying the string $\gamma$ and
the integer $j$, since $S$ and $w_1 \cdots w_j$ are fixed. We shall
denote such an instance by $\langle \gamma,j \rangle$. The proof will
turn on ranking the various instances and proving the result by
induction on the rank.  The rank of the instance $\langle \gamma,j
\rangle$ is computed as the sum of $j$ and the length of a shortest
leftmost derivation of $S\derivestar w_1 \cdots w_j \gamma$.

If the rank is zero, then $j=0$ and $\gamma = S$. Then, we need to show
that $[{} \bullet S,0]$ is generated, which is the case since it is an
axiom of the top-down deduction system.

For the inductive step, let $\langle \gamma,j \rangle$ be an instance
of the lemma of some rank $r > 0$, and assume that the lemma is true
for all instances of smaller rank. Two cases arise.

\begin{description}

\item[Case 1:] $S\derivestar w_1 \cdots w_j \gamma$ in one step.
Therefore, $S \ra w_1 \cdots w_j \gamma$ is a rule of the grammar.
However, since $[{} \bullet S,0]$ is an axiom, by one application of
the prediction rule (predicting the rule $S \ra w_1 \cdots w_j
\gamma$) and $j$ applications of the scanning rule, the item
$[{}\bullet\gamma,j]$ will be generated.

\item[Case 2:] $S\derivestar w_1 \cdots w_j \gamma$ in more than one
step.  Let us assume therefore that $S\derivestar w_1 \cdots w_{j-k} B
\gamma' \Ra w_1 \cdots w_{j} \beta \gamma' $ where $\gamma = \beta
\gamma'$ and $B \ra w_{j-k+1} \cdots w_j \beta$. The instance $\langle
B \gamma', j-k \rangle$ has a strictly smaller rank than $\langle
\gamma, j \rangle$.  Therefore, by the induction hypothesis, the item
$[{}\bullet B \gamma',j-k]$ will be generated.  But then, by
prediction, the item $[{} \bullet w_{j-k+1} \cdots w_j \beta, j-k]$
will be generated and by $k$ applications of the scanning rule, the
item $[\bullet B, j]$ will be generated.

\end{description}

This concludes the proof of the lemma. Completeness of the parser
follows as a corollary of the lemma since if $S \derivestar w_1 \cdots
w_n$, then by the lemma the item $[{} \bullet,n]$ will be generated.

Completeness proofs for the remaining parsing logics discussed in this
paper could be provided in a similar way by relating an appropriate
notion of normal-form derivation for the grammar formalism under
consideration to the item invariants.

\subsection{Pure Bottom-Up Parsing (Shift-Reduce)}

A pure bottom-up algorithm can be specified by such a deduction system
as well.  Here, the items will have the form $[ \alpha \bullet {}, j
]$.  Such an item asserts the dual of the assertion made by the
top-down items, that $ \alpha w_{j+1} \cdots w_n \derivestar w_1
\cdots w_n$ (or, equivalently but less transparently dual, that $
\alpha
\derivestar w_1 \cdots w_j$).  The algorithm is then characterized by
the deduction system shown in Figure~\ref{fig:bu-sys}.  The algorithm
mimics the operation of a nondeterministic shift-reduce parsing
mechanism, where the string of symbols preceding the dot corresponds
to the current parse stack, and the substring starting at the index
$j$ corresponds to the as yet unread input.

\begin{figure}
\begin{center}
\begin{tabular}{ll}

{\bf Item form:}	&	 $[ \alpha \bullet {}, j ]$	\\ \\

{\bf Axioms:}		&	 $[ {} \bullet {}, 0 ]$		\\ \\

{\bf Goals:}		&	 $[ S \bullet {}, n ]$		\\ \\

{\bf Inference Rules:}	\\

{\bf \quad Shift}	&

	\( \oneover{[ \alpha \bullet {}, j ]}
		   {[ \alpha w_{j+1} \bullet {}, j+1 ]}	\)	\\ \\
{\bf \quad Reduce}	&
	\( \oneover{[ \alpha \gamma \bullet {}, j ]}
		   {[ \alpha B \bullet {}, j ]}
			\quad \mbox{ $B \ra \gamma$} \)

\end{tabular}
\end{center}
\caption{The bottom-up shift-reduce deductive parsing system.}
\label{fig:bu-sys}
\end{figure}

The soundness of the inference rules in Figure~\ref{fig:bu-sys} is
easy to see. The antecedent of the shift rule claims that $\alpha
w_{j+1}\cdots w_n \derivestar w_1\cdots w_n$, but that is also what
the consequent claims. For the reduce rule, if $\alpha\gamma
w_{j+1}\cdots w_n \derivestar w_1\cdots w_n$ and $B\ra \gamma$, then
by definition of $\derivestar$ we also have $\alpha B w_{j+1}\cdots
w_n \derivestar w_1\cdots w_n$. As for completeness, it can be proved
by induction on the steps of a reversed rightmost context-free
derivation in a way very similar to the completeness proof of the last
section.

The following derivation shows the operation of the bottom-up rules on
example sentence (\ref{eg-sent}):
\[
\begin{tabular}{rll}
1 & $[ {} \bullet {}, 0 ]$ & \useaxiom \\
2 & $[ \mbox{a} \bullet {}, 1 ]$ & \shift{1} \\
3 & $[ \mbox{\it Det} \bullet {},1 ]$ & \reduce{2} \\
4 & $[ \mbox{\it Det}\; \mbox{program} \bullet {},2 ]$ & \shift{3} \\
5 & $[ \mbox{\it Det}\; N \bullet {},2 ]$ & \reduce{4} \\
6 & $[ \mbox{\it Det}\; N\; \mbox{\it OptRel}\; \bullet {},2 ]$ & \reduce{5} \\
7 & $[ \mbox{\it NP} \bullet {},2 ]$ & \reduce{6} \\
8 & $[ \mbox{\it NP}\; \mbox{halts} \bullet {},3 ]$ & \shift{7} \\
9 & $[ \mbox{\it NP}\; \mbox{\it IV} \bullet {},3 ]$ & \reduce{8} \\
10 & $[ \mbox{\it NP}\; \mbox{\it VP} \bullet {},3 ]$ & \reduce{9} \\
11 & $[ S \bullet {},3 ]$ & \reduce{10}
\end{tabular}
\]
The last item is a goal item, which shows that the sentence is
parsable according to the grammar.

\subsection{Earley's Algorithm}

Stating the algorithms in this way points up the duality of
recursive-descent and shift-reduce parsing in a way that traditional
presentations do not.  The summary presentation in
Figure~\ref{tab:summary} may further illuminate the various
interrelationships.  As we will see, Earley's algorithm \cite{e70} can
then be seen as the natural combination of these two algorithms.

\begin{figure}
\rotate{\vbox{
\begin{center}
\begin{tabular}{l|c|c|c|}
{\bf Algorithm} & {\it Bottom-Up} & {\it Top-Down} & {\it Earley's} \\[2ex]
\hline \hline
&&&\\
{\bf Item form}
        & $[ \alpha \bullet {}, j ]$
        & $[ {} \bullet \beta, j ]$
        & $[ i, A \ra \alpha \bullet \beta, j ]$
                                                        \\
&&&\\
\hline
&&&\\
{\bf Invariant}
        &
        &  $S \derivestar w_1 \cdots w_j \beta$
        &  $S \derivestar w_1 \cdots w_i A \gamma$ \\[2ex]
        &  $\alpha w_{j+1} \cdots w_n \derivestar w_1 \cdots w_n$
        &
        &  $\alpha w_{j+1} \cdots w_n \derivestar w_{i+1} \cdots w_n$
                                                        \\
&&&\\
\hline \hline
&&&\\
{\bf Axioms}
        & $[ {} \bullet {}, 0 ]$
        & $[ {} \bullet S, 0 ]$
        & $[ 0, S' \ra {} \bullet S, 0 ]$
                                                        \\
&&&\\
\hline
&&&\\
{\bf Goals}
        &  $[ S \bullet {}, n ]$
        &  $[ {} \bullet {}, n ]$
        &  $[ 0, S' \ra S \bullet {}, n ]$
                                                        \\
&&&\\
\hline \hline
&&&\\
{\bf Scanning}
        &  $\oneover{[ \alpha \bullet {}, j ]}
                    {[ \alpha w_{j+1} \bullet {}, j+1 ]}$
        &  $\oneover{[ {} \bullet w_{j+1} \beta, j ]}
                    {[ {} \bullet \beta, j+1 ]}$
        &  $\oneover{[ i, A \ra \alpha \bullet w_{j+1} \beta, j ]}
                   {[ i, A \ra \alpha w_{j+1} \bullet \beta, j+1 ]}$
                                                        \\
&&&\\
\hline
&&&\\
{\bf Prediction}
        &
        &  $\oneover{[ {} \bullet B \beta, j ]}
                    {[ {} \bullet \gamma \beta, j ]}
                \quad \mbox{ $B \ra \gamma$}$
        &  $\oneover{[ i, A \ra \alpha \bullet B \beta, j ]}
                   {[ j, B \ra {} \bullet \gamma, j ]}
                \quad \mbox{ $B \ra \gamma$}$
                                                        \\
&&&\\
\hline
&&&\\
{\bf Completion}
        &  $\oneover{[ \alpha \gamma \bullet {}, j ]}
                    {[ \alpha B \bullet {}, j ]}
                \quad \mbox{ $B \ra \gamma$}$
        &
        &  $\oneover{[ i, A \ra \alpha \bullet B \beta, k] \quad
                    [ k, B \ra \gamma \bullet {}, j] }
                   {[ i, A \ra \alpha B \bullet \beta, j]}$
                                                        \\
&&&\\
\hline
\end{tabular}
\end{center}}}
\bigskip
\caption{Summary of parsing algorithms presented as deductive parsing
systems.  (In the axioms and goal items of Earley's algorithm, $S'$
serves as a new nonterminal not in $N$.)}
\label{tab:summary}
\end{figure}

In recursive-descent parsing, we keep a partial sentential form for
the material yet to be parsed, using the dot at the beginning of the
string of symbols to remind us that these symbols come after the point
that we have reached in the recognition process.  In shift-reduce
parsing, we keep a partial sentential form for the material that has
already been parsed, placing a dot at the end of the string to remind
us that these symbols come before the point that we have reached in
the recognition process.  In Earley's algorithm we keep both of these
partial sentential forms, with the dot marking the point somewhere in
the middle where recognition has reached.  The dot thus changes from a
mnemonic to a necessary role.  In addition, Earley's algorithm
localizes the piece of sentential form that is being tracked to that
introduced by a single production.  (Because the first two parsers do
not limit the information stored in an item to only local information,
they are not practical algorithms as stated.  Rather some scheme for
sharing the information among items would be necessary to make them
tractable.)

The items of Earley's algorithm are thus of the form $[ i, A \ra
\alpha \bullet \beta, j ]$ where $\alpha$ and $\beta$ are strings in
$V^*$ and $A \ra \alpha \beta$ is a production of the grammar.  As
was the case for the previous two algorithms, the $j$ index provides
the position in the string that recognition has reached, and the dot
position marks that point in the partial sentential form.  In these
items, however, an extra index $i$ marks the starting position of the
partial sentential form, as we have localized attention to a
single production.  In summary, an item of the form $[ i, A \ra
\alpha \bullet \beta, j ]$ makes the top-down claim that $S \derivestar w_1
\cdots w_i A \gamma$, and the bottom-up claim that $\alpha w_{j+1}
\cdots w_n \derivestar w_{i+1} \cdots w_n$.  The two claims are
connected by the fact that $A \ra \alpha \beta$ is a production in the
grammar.

The algorithm itself is captured by the specification found in
Figure~\ref{tab:summary}. Proofs of soundness and completeness are
somewhat more complex than those for the pure top-down and bottom-up
cases shown above, and are directly related to the corresponding
proofs for Earley's original algorithm \cite{e70}.

The following derivation, again for sentence
(\ref{eg-sent}), illustrates the operation of the Earley inference
rules:
\[
\begin{tabular}{rll}
1 & $[0, S'\ra {} \bullet S, 0 ]$ & \useaxiom \\
2 & $[0, S \ra  {}\bullet \mbox{\it NP}\; \mbox{\it VP}, 0 ]$ & \predict{1}\\
3 & $[0, \mbox{\it NP}\ra  {}\bullet  \mbox{\it Det}\; N\; \mbox{\it OptRel},0
]$ & \predict{2}\\
4 & $[0, \mbox{\it Det} \ra  {}\bullet  \mbox{a},0 ]$ & \predict{3}\\
5 & $[0, \mbox{\it Det} \ra \mbox{a}\bullet {},1]$ & \scan{4}\\
6 & $[0, \mbox{\it NP}\ra \mbox{\it Det} \bullet N\; \mbox{\it OptRel},1 ]$ &
\complete{3}{5} \\
7 & $[1, N \ra {} \bullet \mbox{program}, 1]$ & \predict{6} \\
8 & $[1, N \ra \mbox{program} \bullet {}, 2]$ & \scan{7} \\
9 & $[0, \mbox{\it NP}\ra \mbox{\it Det}\; N \bullet \mbox{\it OptRel},2 ]$ &
\complete{6}{8}\\
10 & $[2, \mbox{\it OptRel}\ra {}\bullet {}, 2]$ & \predict{9} \\
11 & $[0, \mbox{\it NP}\ra \mbox{\it Det}\; N \; \mbox{\it OptRel}\; \bullet
{},2 ]$ & \complete{9}{10} \\
12 & $[0, S \ra \mbox{\it NP}\bullet \mbox{\it VP}, 2]$ & \complete{2}{11} \\
13 & $[2, \mbox{\it VP}\ra {} \bullet \mbox{\it IV},2 ]$ & \predict{12} \\
14 & $[2, \mbox{\it IV}\ra {} \bullet \mbox{halts},2 ]$ & \predict{13} \\
15 & $[2, \mbox{\it IV}\ra \mbox{halts}\bullet {},3 ]$ & \scan{14} \\
16 & $[2, \mbox{\it VP}\ra \mbox{\it IV}\bullet {},3 ]$ & \complete{13}{15} \\
17 & $[0, S \ra \mbox{\it NP}\; \mbox{\it VP}\bullet {}, 3]$ &
\complete{12}{16} \\
18 & $[0, S'\ra S \bullet {}, 3 ]$ & \complete{1}{17}
\end{tabular}
\]
The last item is again a goal item, so we have an Earley derivation of the
grammaticality of the given sentence.

\section{Deductive Parsing for Other Formalisms}
\label{sec:others}

The methods (and implementation) that we developed have also been used
for rapid prototyping and experimentation with parsing algorithms for
grammatical frameworks other than context-free grammars.  They can be
naturally extended to handle augmented phrase-structure formalisms
such as logic grammar and constraint-based formalisms.  They have been
used in the development and testing of algorithms for parsing
categorial grammars, tree-adjoining grammars, and lexicalized
context-free grammars.  In this section, we discuss these and other
extensions.

\subsection{Augmented Phrase-Structure Formalisms}
\label{sec:augmented}

It is straightforward to see that the three deduction systems just
presented can be extended to constraint-based grammar formalisms with
a context-free backbone. The basis for this extension goes back to
metamorphosis grammars \cite{Colmerauer:mgs} and definite-clause
grammars (DCG) \cite{Pereira+Warren:DCGs}. In those formalisms,
grammar symbols are first-order terms, which can be understood as
abbreviations for the sets of all their ground instances. Then an
inference rule can also be seen as an abbreviation for all of its
ground instances, with the metagrammatical variables in the rule
consistently instantiated to ground terms. Computationally, however,
such instances are generated lazily by accumulating the consistency
requirements for the instantiation of inference rules as a conjunction
of equality constraints and maintaining that conjunction in normal
form --- sets of variable substitutions --- by unification. (This is
directly related to the use of unification to avoid ``guessing''
instances in the rules of existential generalization and universal
instantiation in a natural-deduction presentation of first-order
logic).

We can move beyond first-order terms to general constraint-based
grammar formalisms \cite{Shieber:inference,Carpenter:logic} by taking
the above constraint interpretation of inference rules as basic. More
explicitly, a rule such as Earley completion
\[
\oneover{[ i, A \ra \alpha \bullet B \beta, k] \quad
                    [ k, B \ra \gamma \bullet {}, j] }
                   {[ i, A \ra \alpha B \bullet \beta, j]}
\]
is interpreted as shorthand for the constrained rule:
\[
\oneover{[ i, A \ra \alpha \bullet B \beta, k] \quad
                    [ k, B' \ra \gamma \bullet {}, j] }
                   {[ i, A' \ra \alpha B'' \bullet \beta,
j]}\quad\mbox{ $A=A'$ and $B=B'$ and $B=B''$}
\]
When such a rule is applied, the three constraints it depends on are
conjoined with the constraints for the current derivation.
In the particular case of first-order terms and
antecedent-to-consequent rule application, completion can be given
more explicitly as
\[
\oneover{[ i, A \ra \alpha \bullet B \beta, k] \quad
                    [ k, B' \ra \gamma \bullet {}, j] }
                   {[ i, \sigma(A \ra \alpha B \bullet \beta),
j]}\quad\mbox{ $\sigma = \mbox{\rm mgu}(B,B')$} \qquad .
\]
where $mgu(B, B')$ is the most general unifier of the terms $B$ and
$B'$.  This is the interpretation implemented by the deduction
procedure described in the next section.

The move to constraint-based formalisms raises termination problems in
proof construction that did not arise in the context-free case. In the
general case, this is inevitable, because a formalism like DCG
\cite{Pereira+Warren:DCGs} or PATR-II \cite{Shieber:criteria} has
Turing-machine power. However, even if constraints are imposed on the
context-free backbone of the grammar productions to guarantee decidability,
such as {\em offline parsability}
\cite{Bresnan+Kaplan:lfg,Pereira+Warren:ED,Shieber:inference}, the
prediction rules for the top-down and Earley systems are problematic.
The difficulty is that prediction can feed on its own results to build
unboundedly large items. For example, consider the DCG
\[
\begin{tabular}{l}
$s \ra r(0,N)$ \\
$r(X,N) \ra r(s(X),N)\; b$ \\
$r(N,N) \ra a$
\end{tabular}
\]
It is clear that this grammar accepts strings of the form $ab^n$ with
the variable $N$ being instantiated to the unary (successor)
representation of $n$. It is also clear that the bottom-up inference
rules will have no difficulty in deriving the analysis of any input
string. However, Earley prediction from the item $[0,s \ra {} \bullet
r(0,N), 0]$ will generate an infinite succession of items:
\[
\begin{tabular}{l}
$[0, s \ra {}\bullet r(0,N), 0]$ \\
$[0, r(0 ,N) \ra {} \bullet  r(s(0),N)\; b, 0]$ \\
$[0, r(s(0) ,N) \ra {} \bullet  r(s(s(0)),N)\; b, 0]$ \\
$[0, r(s(s(0)) ,N) \ra {} \bullet  r(s(s(s(0))),N)\; b, 0]$ \\
$\cdots$
\end{tabular}
\]
This problem can be solved in the case of the Earley inference rules
by observing that prediction is just used to narrow the number of
items to be considered by scanning and completion, by maintaining the
top-down invariant $S \derivestar w_1 \cdots w_i A \gamma$. But this
invariant is not required for soundness or completeness, since the
bottom-up invariant is sufficient to guarantee that items represent
well-formed substrings of the input.  The only purpose of the top-down
invariant is to minimize the number of completions that are actually
attempted. Thus the only indispensable role of prediction is to make
available appropriate instances of the grammar productions. Therefore, any
relaxation of prediction that makes available items of which all the
items predicted by the original prediction rule are instances will not
affect soundness or completeness of the rules. More precisely, it must
be the case that any
item $[i, B \ra {} \bullet \gamma, i]$ that the original
prediction rule would create is an instance  of some item $[i, B' \ra {}
\bullet \gamma', i]$ created by the relaxed prediction rule.
A relaxed prediction rule will create no more items
than the original predictor, and in fact may create far fewer. In
particular, repeated prediction may terminate in cases like the one
described above. For example, if the prediction rule applied to
$[i,A\ra\alpha \bullet B' \beta,j]$ yields $[i,\sigma(B\ra {} \bullet
\gamma),i]$ where $\sigma=\mbox{\rm mgu}(B,B')$, a relaxed prediction
rule might yield $[i,\sigma'(B\ra {} \bullet
\gamma),i]$, where $\sigma'$ is a less specific substitution than
$\sigma$ chosen so that only a finite number of instances of
$[i,B\ra{}\bullet\gamma,i]$ are ever generated. A similar notion for
general constraint grammars is called restriction
\cite{Shieber:restriction,Shieber:inference}, and a related technique
has been used in partial evaluation of logic programs
\cite{Sato+Tamaki:patterns}.

\label{sec:offline} The problem with the DCG above can be seen as
following from the computation of derivation-specific information in
the arguments to the nonterminals.  However, applications frequently
require construction of the derivation for a string (or similar
information), perhaps for the purpose of further processing.  It is
simple enough to augment the inference rules to include with each item
a derivation.  For the Earley deduction system, the items would
include a fourth component whose value is a sequence of derivation
trees, nodes labeled by productions of the grammar, one derivation
tree for each element of the right-hand side of the item before the
dot.  The inference rules would be modified as shown in
Figure~\ref{fig:earley-mod}.  The system makes use of a function
$tree$ that takes a node label $l$ (a production in the grammar) and a
sequence of derivation trees $D$ and forms a tree whose root is
labeled by $l$ and whose children are the trees in $D$ in order.

\begin{figure}
\begin{center}
\begin{tabular}{ll}

{\bf Item form:}
	& $[ i, A \alpha \bullet \beta, j, D ]$		\\ \\

{\bf Axioms:}
        & $[ 0, S' \ra {} \bullet S, 0, \seq{} ]$
                                                        \\ \\
{\bf Goals:}
        &  $[ 0, S' \ra S \bullet {}, n, D ]$
                                                        \\ \\

{\bf Inference rules:}\\

{\bf \quad Scanning}
        &  $\oneover{[ i, A \ra \alpha \bullet w_{j+1} \beta, j, D ]}
                   {[ i, A \ra \alpha w_{j+1} \bullet \beta, j+1, D ]}$
                                                        \\  \\
{\bf \quad Prediction}
        &  $\oneover{[ i, A \ra \alpha \bullet B \beta, j, D ]}
                   {[ j, B \ra {} \bullet \gamma, j, \seq{} ]}
                \quad \mbox{ $B \ra \gamma$}$
                                                        \\  \\
{\bf \quad Completion}
        &  $\oneover{[ i, A \ra \alpha \bullet B \beta, k, D_1 ] \quad
                    [ k, B \ra \gamma \bullet {}, j, D_2 ] }
                   {[ i, A \ra \alpha B \bullet \beta, j,
			D_1 \cup tree(B \ra \gamma, D_2) ]}$
\end{tabular}
\end{center}
\caption{The Earley deductive parsing system modified to generate
derivation trees.}
\label{fig:earley-mod}
\end{figure}

Of course, use of such rules makes the caching of lemmas essentially
useless, as lemmas derived in different ways are never identical.
Appropriate methods of implementation that circumvent this problem are
discussed in Section~\ref{sec:imp-offline}.

\subsection{Combinatory Categorial Grammars} \label{sec:ccg}

A combinatory categorial grammar \cite{ades-steedman} consists of two
parts: (1) a lexicon that maps words to sets of categories; (2) rules
for combining categories into other categories.

Categories are built from atomic categories and two binary operators:
forward slash ($\fw$) and backward slash ($\bk$). Informally speaking,
words having categories of the form $X \fw Y$, $X \bk Y$, $(W \fw X)
\fw Y$ etc.  are to be thought of as functions over $Y$'s. Thus the
category $S \bk \NP$ of intransitive verbs should be interpreted as a
function from noun phrases ($\NP$) to sentences ($S$). In addition,
the direction of the slash (forward as in $X \fw Y$ or backward as in
$X \bk Y$) specifies where the argument must be found, immediately to
the right for $ \fw $ or immediately to the left for $\bk$.

For example, a CCG lexicon may assign the category $ S \bk \NP$ to an
intransitive verb (as the word {\em sleeps}). $ S \bk \NP$ identifies the word
({\em sleeps}) as combining with a (subject) noun phrase ($\NP$) to yield a
sentence ($S$).  The back slash ($\bk$) indicates that the subject must be
found immediately to the left of the verb.  The forward slash $ \fw $ would
have
indicated that the argument must be found immediately to the right of the
verb.

More formally, categories are defined inductively as
follows:\footnote{The notation for backward slash used in this paper
is consistent with one defined by Ades and Steedman
\shortcite{ades-steedman}: $X \bk Y$ is interpreted as a function from $Y$s
to $X$s.  Although this notation has been adopted by the majority of
combinatory categorial grammarians, other frameworks
\cite{Lambek:sentstruct} have adopted the opposite interpretation for
$X \bk Y$: a function from $X$s to $Y$s.} Given a set of nonterminals,

\begin{itemize}
\item Nonterminal symbols are categories.
\item If $c_1$ and $c_2$ are categories, then $(c_1 \fw c_2)$ and
$(c_1 \bk c_2)$ are categories.
\end{itemize}

\begin{figure}
\begin{center}
\begin{tabular}{ll}
{\it Word} & {\it Category} \\
\hline
John & $\NP$\\
bananas & $\NP$\\
likes &  $ (S \bk \NP)  \fw  \NP$\\
really & $(S \bk \NP)  \fw  (S \bk \NP)$
\end{tabular}
\end{center}
\caption{An example CCG lexicon.}
\label{fig:ccgtoy}
\end{figure}

The lexicon is defined as a mapping $f$ from words to finite sets of
categories. Figure~\ref{fig:ccgtoy} is an example of a CCG lexicon.
In this lexicon, $likes$ is encoded as a transitive verb $(S \bk \NP)
\fw \NP$, yielding a sentence ($S$) when a noun phrase ($\NP$) object
is found to its right and when a noun phrase subject ($\NP$) is then
found to its left.

Categories can be combined by a finite set of rules that fall in two
classes: application and composition.

Application allows the simple combination of a function with an
argument to its right (forward application) or to its left (backward
application). For example, the sequence $(S \bk \NP) \fw \NP \quad
\NP$ can be reduced to $S \bk \NP$ by applying the forward application
rule. Similarly, the sequence $\NP \quad S \bk \NP$ can be reduced to
$S$ by applying the backward application rule.

Composition allows to combine two categories in a similar fashion as
functional composition. For example, forward composition combines two
categories of the form $X \fw Y \ Y \fw Z$ to another category $X \fw
Z$.  The rule gives the appearance of ``canceling'' $Y$, as if the two
categories were numerical fractions undergoing multiplication. This
rule corresponds to the fundamental operation of ``composing'' the two
functions, the function $X \fw Y$ from $Y$ to $X$ and the function $Y
\fw Z$ from $Z$ to $Y$.

The rules of composition can be specified formally as productions, but
unlike the productions of a CFG, these productions are universal over
all CCGs.  In order to reduce the number of cases, we will use a
vertical bar $|$ as an instance of a forward or backward slash, $ \fw
$ or $\bk$.  Instances of $|$ in left- and right-hand sides of a
single production should be interpreted as representing slashes of the
same direction.  The symbols $X$, $Y$ and $Z$ are to be read as
variables which match any category.

\begin{flushleft}
\begin{tabular}{lrcl}
{\bf Forward application:} &
\ccgrule{X \fw Y \quad Y}
           {X}
\\
{\bf Backward application:} &
\ccgrule{Y \quad X \bk Y}
	   {X}
\\
{\bf Forward composition:} &
\ccgrule{X  \fw  Y \quad Y | Z}
	   {X | Z}
\\
{\bf Backward composition:} &
\ccgrule{Y | Z \quad X \bk Y}
	   {X|Z}
\end{tabular}
\end{flushleft}

\noindent A string of words is accepted by a CCG, if a specified nonterminal
symbol (usually $S$) derives a string of categories that is an image
of the string of words under the mapping $f$.

A bottom-up algorithm --- essentially the CYK algorithm instantiated
for these productions --- can be easily specified for CCGs. Given a
CCG, and a string $w = w_1 \cdots w_n$ to be parsed, we will consider
a logic with items of the form $[ X, i, j ]$ where $X$ is a category
and $i$ and $j$ are integers ranging from $0$ to $n$.  Such an item,
asserts that the substring of the string $w$ from the $i+1$-th element
up to the $j$-th element can be reduced to the category $X$.  The
required proof rules for this logic are given in Figure \ref{fig:ccg}.

\begin{figure}
\begin{center}
\begin{tabular}{ll}

{\bf Item form:}	&	$[ X, i, j ]$	\\ \\

{\bf Axioms:}		&	$[ X, i, i+1 ]$ \qquad  where $X \in f(w_{i+1})$ \\ \\

{\bf Goals:}		&	 $[ S, 0, n ]$\\ \\

{\bf Inference rules:}	&	\\

{\bf \quad Forward Application}	&
	\( \oneover{[X \fw Y,i,j] \quad [Y,j,k]}
		   {[X,i,j]} \)	\\ \\
{\bf \quad Backward Application}&
	\( \oneover{[Y,i,j] \quad [X \bk Y,j,k]}
		   {[X,i,k]} \)	\\ \\
{\bf \quad Forward Composition 1}&
	\( \oneover{[X \fw Y,i,j] \quad [Y \fw Z,j,k]}
		   {[X \fw Z,i,k]} \)	\\ \\
{\bf \quad Forward Composition 2}&
	\( \oneover{[X \fw Y,i,j] \quad [Y \bk Z,j,k]}
		   {[X \bk Z,i,k]} \)	\\ \\
{\bf \quad Backward Composition 1}&
	\( \oneover{[Y \fw Z,i,j] \quad [X \bk Y, j, k]}
		   {[X \fw Z,i,k]} \)	\\ \\
{\bf \quad Backward Composition 2}&
	\( \oneover{[Y \bk Z,i,j] \quad [X \bk Y, j, k]}
		   {[X \bk Z,i,k]} \)

\end{tabular}
\end{center}
\caption{The CCG deductive parsing system.}
\label{fig:ccg}
\end{figure}

To illustrate the operations, we will use the lexicon in
Figure~\ref{fig:ccgtoy} to combine the string

\begin{equation}
\mbox{John really likes bananas}
\label{eg-sent-ccg}
\end{equation}

Among other ways, the sentence can be proved as follows:

\[
\begin{tabular}{rll}
1 & $[\NP,0,1]$ & \useaxiom \\
2 & $[(S \bk \NP)  \fw  (S \bk \NP),1,2]$ & \useaxiom \\
3 & $[(S \bk \NP)  \fw  \NP,2,3]$ & \useaxiom \\
4 & $[(S \bk \NP)  \fw  \NP,1,3]$ & \userule{forward composition}{2 and 3} \\
5 & $[\NP,3,4]$ & \useaxiom \\
6 & $[(S \bk \NP),1,4]$ & \userule{forward application}{4 and 5} \\
7 & $[S,0,4]$ & \userule{backward application}{1 and 6}
\end{tabular}
\]

Other extensions of CCG (such as generalized composition and
coordination) can be easily implemented using such deduction parsing
methods.

\subsection{Tree-Adjoining Grammars and Related Formalisms}

The formalism of tree-adjoining grammars (TAG) \cite{jlt75,j83} is a
tree-generating system in which trees are combined by an operation of
adjunction rather than the substitution operation of context-free
grammars.\footnote{Most practical variants of TAG include both
adjunction and substitution, but for purposes of exposition we
restrict our attention to adjunction alone, since substitution is
formally dispensible and its implementation in parsing systems such as
we describe is very much like the context-free operation.  Similarly,
we do not address other issues such as adjoining constraints and extended
derivations.  Discussion of those can be found elsewhere
\cite{schabes-ci94,ss92a}.} The increased expressive power of
adjunction allows important natural-language phenomena such as
long-distance dependencies to be expressed {\em locally} in the
grammar, that is, within the relevant lexical entries, rather than by
many specialized context-free rules
\cite{kj85}.

\begin{figure}
\begin{center}
\mbox{\psfig{figure=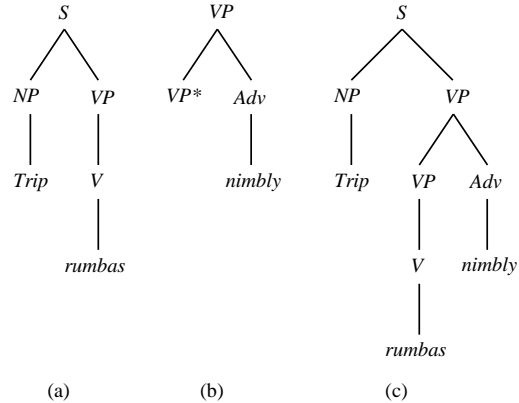,scale=80}}
\end{center}
\caption{An example tree-adjoining grammar consisting of one initial
tree (a), and one auxiliary tree (b). These trees can be used to form
the derived tree (c) for the sentence ``Trip rumbas nimbly.''  (In an
actual English grammar, the tree depicted in (a) would not be an
elementary tree, but itself derived from two trees, one for each
lexical item, by a substitution operation.)}
\label{fig:example}
\end{figure}

A tree-adjoining grammar consists of a set of {\em elementary trees}
of two types: {\em initial trees} and {\em auxiliary trees}.  An
initial tree is complete in the sense that its frontier includes only
terminal symbols.  An example is given in Figure~\ref{fig:example}(a).
An auxiliary tree is incomplete; it has a single node on the frontier,
the {\em foot node}, labeled by the same nonterminal as the root.
Figure~\ref{fig:example}(b) provides an example.  (By convention, foot
nodes are redundantly marked by a diacritic asterisk ($\ast$) as in
the figure.)

\begin{figure}
\begin{center}
\mbox{\psfig{figure=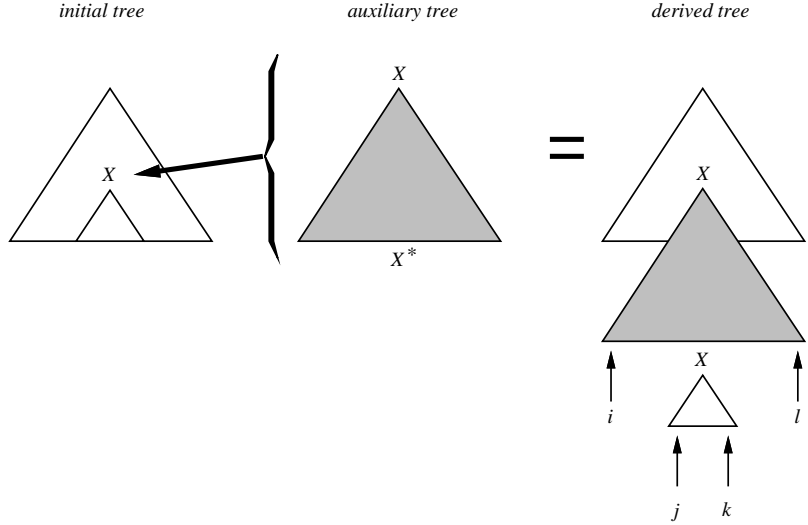,scale=80}}
\end{center}
\caption{The operation of adjunction.  The auxiliary tree is spliced
into the initial tree to yield the derived tree at right.}
\label{fig:adjoin}
\end{figure}

Although auxiliary trees do not themselves constitute complete
grammatical structures, they participate in the construction of
complete trees through the adjunction operation.  Adjunction of an
auxiliary tree into an initial tree is depicted in
Figure~\ref{fig:adjoin}.  The operation inserts a copy of an auxiliary
tree into another tree in place of an interior node that has the same
label as the root and foot nodes of the auxiliary tree.  The subtree
that was previously connected to the interior node is reconnected to
the foot node of the copy of the auxiliary tree.  For example, the
auxiliary tree in Figure~\ref{fig:example}(b) can be adjoined at the
{\em VP} node of the initial tree in Figure~\ref{fig:example}(a) to
form the derived tree in Figure~\ref{fig:example}(c).  Adjunction in
effect supports a form of string wrapping and is therefore more
powerful than the substitution operation of context-free grammars.

A tree-adjoining grammar can be specified as a quintuple $G = \seq{N,
\Sigma, I, A, S}$, where $N$ is the set of nonterminals including the
start symbol $S$, $\Sigma$ is the disjoint set of terminal symbols,
$I$ is the set of initial trees, and $A$ is the set of auxiliary
trees.

To describe adjunction and TAG derivations, we need notation to refer
to tree nodes, their labels, and the subtrees they define.  Every node
in a tree $\alpha$ can be specified by its {\em address}, a sequence
of positive integers defined inductively as follows: the empty
sequence $\epsilon$ is the address of the root node, and $\address
\cdot k$ is the address of the $k$-th child of the node at address
$\address$. $\treefoot{\alpha}$ is defined as the address of the foot
node of the tree $\alpha$ if there is one; otherwise
$\treefoot{\alpha}$ is undefined.

We denote by $\anode{\alpha}{\address}$ the node of $\alpha$ at
address $\address$, and by $\subtree{\alpha}{\address}$ the subtree of
$\alpha$ rooted at $\anode{\alpha}{\address}$. The grammar symbol that
labels node $\nu$ is denoted by $\nodelab{\nu}$.  Given an elementary
tree node $\nu$, $\adjoinable{\nu}$ is defined as the set of auxiliary
trees that can be adjoined at node $\nu$.\footnote{For TAGs with no
constraints on adjunction (for instance, as defined here),
$\adjoinable{\nu}$ is just the set of elementary auxiliary trees whose
root node is labeled by $\nodelab{\nu}$.  When other adjoining
constraints are allowed, as is standard, they can be incorporated
through a revised definition of $Adj$.}

Finally, we denote by $\alpha[\beta_1\mapsto\address_1, \ldots,
\beta_k\mapsto\address_k]$ the result of adjoining the trees  $\beta_1,
\ldots, \beta_k$ at distinct addresses $\address_1,
\ldots, \address_k$ in the tree $\alpha$.

The set of trees $D(G)$ derived by a TAG $G$ can be defined
inductively.  $D(G)$ is the smallest set of trees such that

\begin{enumerate}

\item $I \cup A \subseteq D(G)$, that is, all elementary trees are
derivable, and

\item Define $D(\alpha,G)$ to be the set of all trees derivable as
$\alpha[\beta_1\mapsto\address_1, \ldots, \beta_k\mapsto\address_k]$
where $\beta_1, \ldots, \beta_k \in D(G)$ and $\address_1, \ldots,
\address_k$ are distinct addresses in $\alpha$.  Then, for
all elementary trees $\alpha \in I \cup A$, $D(\alpha,G) \subseteq D(G)$.
Obviously, if $\alpha$ is an initial tree, the tree thus derived will
have no foot node, and if $\alpha$ is an auxiliary tree, the derived
tree will have a foot node.

\end{enumerate}

\noindent The valid derivations in a TAG are the trees in
$D(\alpha_S,G)$ where $\alpha_S$ is an initial tree whose root is
labeled with the start symbol $S$.

Parsers for TAG can be described just as those for CFG, as deduction
systems.  The parser we present here is a variant of the CYK algorithm
extended for TAGs, similar, though not identical, to that of
Vijay-Shanker \shortcite{v87}.  We chose it for expository reasons: it is
by far the simplest TAG parsing algorithm, in part because it is
restricted to TAGs in which elementary trees are at most binary
branching, but primarily because it is purely a bottom-up system; no
prediction is performed.  Despite its simplicity, the algorithm must
handle the increased generative capacity of TAGs over that of
context-free grammars.  Consequently, the worst case complexity for
the parser we present is worse than for CFGs --- $O(n^6)$ time for a
sentence of length $n$.

The present algorithm uses a {\em dotted tree} to track the progress
of parsing.  A dotted tree is an elementary tree of the grammar with a
dot adjacent to one of the nodes in the tree.  The dot itself may be
in one of two positions relative to the specified node: above or
below.  A dotted tree is thus specified as an elementary tree
$\alpha$, an address $\address$ in that tree, and a marker to specify
the position of the dot relative to the node. We will use the notation
$\nu\posa$ and $\nu\posb$ for dotted trees with the dot above and
below node $\nu$, respectively.\footnote{Although both this algorithm
and Earley's use a dot in items to distinguish the progress of a
parse, they are used in quite distinct ways.  The dot of Earley's
algorithm tracks the left-to-right progress of the parse among
siblings.  The dot of the CYK TAG parser tracks the
pre-/post-adjunction status of a single node.  For this reason, when
generalizing Earley's algorithm to TAG parsing \cite{schabes-ci94},
four dot positions are used to simultaneously track
pre-/post-adjunction and before/after node left-to-right progress.}

In order to track the portion of the string covered by the production
up to the dot position, the CYK algorithm makes use of two indices.
In a dotted tree, however, there is a further complication in that the
elementary tree may contain a foot node so that the string covered by
the elementary tree proper has a gap where the foot node occurs.
Thus, in general, four indices must be maintained: two ($i$ and $l$ in
Figure~\ref{fig:adjoin}) to specify the left edge of the auxiliary
tree and the right edge of the parsed portion (up to the dot position)
of the auxiliary tree, and two more ($j$ and $k$) to specify the
substring dominated by the foot node.

The parser therefore consists of inference rules over items of the
following forms:
\( [\nu\posa, i, j, k, l] \) and
\( [\nu\posb, i, j, k, l] \), where

\begin{itemize}

	\item $\nu$ is a node in an elementary tree,

	\item $i,j,k,l$ are indices of positions in the input string
$w_1 \cdots w_n$ ranging over $\{0,\cdots,n\} \cup \{\_\}$, where $\_$
indicates that the corresponding index is not used in that particular
item.

\end{itemize}

An item of the form $[\anode{\alpha}{\address}\posa, i, \_, \_, l]$
specifies that there is a tree $T \in D(\alpha/\address,G)$, with no
foot node, such that the fringe of $T$ is the string $w_{i+1} \cdots
w_l$. An item of the form $[\anode{\alpha}{\address}\posa, i, j, k,
l]$ specifies that there is a tree $T \in D(\alpha/\address,G)$, with
a foot node, such that the fringe of $T$ is the string $w_{i+1} \cdots
w_j\; \nodelab{\treefoot{T}}\; w_{k+1} \cdots w_l$.  The invariants
for $[\anode{\alpha}{\address}\posb, i, \_, \_, l]$ and
$[\anode{\alpha}{\address}\posb, i, j, k, l]$ are similar, except that
the derivation of $T$ must not involve adjunction at node
$\anode{\alpha}{\address}$.

\begin{figure}
\centerline{
\mbox{\psfig{figure=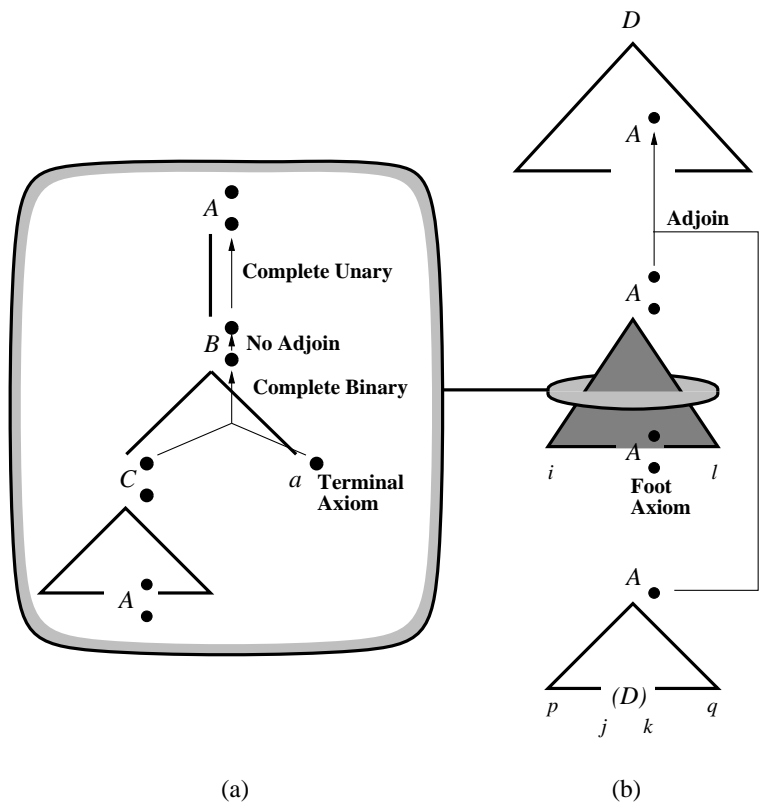,scale=50}}}
\caption{Examples of dot movement in the CYK tree traversal implicit
in the TAG parsing algorithm.}
\label{fig:trav}
\end{figure}

The algorithm preserves this invariant while traversing the derived
tree from bottom to top, starting with items corresponding to the
string symbols themselves, which follow from the axioms
\[	[ \nu\posa, i, \_, \_ , i+1]
		\qquad \nodelab{\nu} = w_{i+1}
\]
combining completed subtrees into larger ones, and combining subtrees
before adjunction (with dot below) and derived auxiliary trees to form
subtrees after adjunction (with dot above).  Figure~\ref{fig:trav}
depicts the movement of the dot from bottom to top as parsing
proceeds.  In Figure~\ref{fig:trav}(a), the basic rules of dot
movement not involving adjunction are shown, including the axiom for
terminal symbols, the combination of two subchildren of a binary tree
and one child of a unary subtree, and the movement corresponding to
the absence of an adjunction at a node.  These are exactly the rules
that would be used in parsing within a single elementary tree.
Figure~\ref{fig:trav}(b) displays the rules involved in parsing an
adjunction of one tree into another.

These dot movement rules are exactly the inference rules of the TAG
CYK deductive parsing system, presented in full in
Figure~\ref{fig:tag-parse}.  In order to reduce the number of cases,
we define the notation $i \cup j$ for two indices $i$ and $j$ as
follows:

\[
i \cup j =
	\left\{
	\begin{array}{ll}
	i		& j = \_\\
	j 		& i = \_\\
	i 		& i = j \\
	\mbox{\em undefined} 	& \mbox{\em otherwise}
	\end{array}
	\right.
\]

\begin{figure}
\begin{center}
\begin{tabular}{ll}

{\bf Item form:}&
	$[ \nu\posa, i, j, k , l ]$ \\
	&$[ \nu\posb, i, j, k , l ]$ \\  \\

{\bf Axioms:} \\
{\bf \quad Terminal Axiom}
& 	$[ \nu\posa, i, \_, \_ , i+1]$
		\qquad $\nodelab{\nu} = w_{i+1}$\\ \\

{\bf \quad Empty String Axiom}
&	$[ \nu\posa, i, \_, \_ , i]$
		\qquad $\nodelab{\nu} = \epsilon$\\ \\

{\bf \quad Foot Axiom}
&	$[\anode{\beta}{\treefoot{\beta}}\posb,p,p,q,q]$
		\qquad $\beta \in A$\\ \\

{\bf Goals:}&
	$[ \anode{\alpha}{\epsilon}\posa, 0, \_, \_ , n]$
		\qquad $\alpha \in I$ and $\nodelab{\anode{\alpha}{\epsilon}} = S$\\ \\

{\bf Inference Rules:} \\

{\bf \quad Complete Unary}	&
	\( \oneover{[\anode{\alpha}{(\address\cdot 1)}\posa,i,j,k,l]}
		   {[\anode{\alpha}{\address}\posb,i,j,k,l]}  \)
		\qquad $\anode{\alpha}{(\address \cdot 2)}$ undefined \\ \\

{\bf \quad Complete Binary}	&
	\( \oneover{[\anode{\alpha}{(\address\cdot 1)}\posa,i,j,k,l] \quad
		    [\anode{\alpha}{(\address\cdot 2)}\posa,l,j',k',m]}
		   {[\anode{\alpha}{\address}\posb,i,j \cup j',k \cup k',m]}\) \\ \\

{\bf \quad No Adjoin}		&
	\( \oneover{[\nu\posb,i,j,k,l]}
		   {[\nu\posa,i,j,k,l]}  \)  \\ \\

{\bf \quad Adjoin}		&
	\( \oneover{[\anode{\beta}{\epsilon}\posa,i,p,q,l] \quad
		    [\nu\posb,p,j,k,q]}
		   {[\nu\posa,i,j,k,l]}  \)
		\qquad $\beta \in \adjoinable{\nu}$

\end{tabular}
\end{center}
\caption{The CYK deductive parsing system for tree-adjoining grammars.}
\label{fig:tag-parse}
\end{figure}

Although this parser works in time $O(n^6)$ --- the Adjoin rule with
its six independent indices is the step that accounts for this
complexity --- and its average behavior may be better, it is in
practice too inefficient for practical use for two reasons.  First, an
attempt is made to parse all auxiliary trees starting bottom-up from
the foot node, regardless of whether the substring between the foot
indices actually can be parsed in an appropriate manner.  This problem
can be alleviated, as suggested by Vijay-Shanker and Weir \shortcite{vw93},
by replacing the Foot Axiom with a Complete Foot rule that generates
the item $[\anode{\beta}{\treefoot{\beta}}\posb,p,p,q,q]$ only if
there is an item $[\nu\posb,p,j,k,q]$ where $\beta \in
\adjoinable{\nu}$, i.e.,
\begin{flushleft}
\begin{tabular}{ll}
{\bf  \quad Complete Foot}	&
	\( \oneover{[\nu\posb,p,j,k,q]}
		   {[\anode{\beta}{\treefoot{\beta}}\posb,p,p,q,q]}\)
		\qquad $\beta \in \adjoinable{\nu}$
\end{tabular}
\end{flushleft}
This complicates the invariant considerably, but makes auxiliary tree
parsing much more goal-directed.  Second, because of the lack of
top-down prediction, attempts are made to parse elementary trees that
are not consistent with the left context.  Predictive parsers for TAG
can be, and have been, described as deductive systems.  For instance,
Schabes \shortcite{schabes-ci94} provides a detailed explanation for a
predictive left-to-right parser for TAG inspired by the techniques of
Earley's algorithm.  Its worst-case complexity is $O(n^6)$ as well,
but its average complexity on English grammar is well superior to its
worst case, and also to the CYK TAG parser.  A parsing system based on
this algorithm is currently being used in the development of a large
English tree-adjoining grammar at the University of Pennsylvania
\cite{xtag92}.

Many other formalisms related to tree-adjoining grammars have been
proposed, and the deductive parsing approach is applicable to these as
well.  For instance, as part of an investigation of the precise
definition of TAG derivation, Schabes and Shieber describe a
compilation of tree-adjoining grammars to linear indexed grammars,
together with an efficient algorithm, stated as deduction system
for recognition and parsing according to the compiled grammar
\cite{ss92a}.  A prototype of this parser has been implemented using
the deduction engine described here.  (In fact, it was as an aid to
testing this algorithm, with its eight inference rules each with as
many as three antecedent items, that the deductive parsing
meta-interpreter was first built.)

Schabes and Waters \shortcite{schabes93} suggest the use of a restricted
form of TAG in which the foot node of an auxiliary tree can occur only
at the left or right edge of the tree.  Since the portion of string
dominated by an auxiliary tree is contiguous under this constraint,
only two indices are required to track the parsing of an auxiliary
tree adjunction.  Consequently, the formalism can generate only
context-free languages and can be parsed in cubic time.  The resulting
system, called lexicalized context-free grammar (LCFG), is a
compromise between the parsing efficiency of context-free grammar and
the elegance and lexical sensitivity of tree-adjoining grammar. The
deductive parsing meta-interpreter has also been used for rapid
prototyping of an Earley-style parser for LCFG \cite{sw93-tr-04}.

\subsection{Inadequacy for Sequent Calculi}

All the parsing logics discussed here have been presented in a
natural-deduction format that can be implemented directly by bottom-up
execution. However, important parsing logics, in particular the Lambek
calculus \cite{Lambek:sentstruct,Moortgat:categorial}, are better
presented in a sequent-calculus format. The main reason for this is
that those systems use nonatomic formulas that represent concurrent or
hypothetical analyses. For instance, if for arbitrary $u$ with
category $B$ we conclude that $vu$ has category $A$, then in the
Lambek calculus we can conclude that $v$ has category $A \fw B$.

The main difficulty with applying our techniques to sequent systems is
that computationally they are designed to be used in a top-down
direction. For instance, the rule used for the hypothetical analysis
above has the form:
\begin{equation}
\oneover{\Gamma B \vdash A}{\Gamma \vdash A \fw B}\label{right-slash}
\end{equation}
It is reasonable to use this rule in a goal-directed fashion
(consequent to antecedent) to show $\Gamma \vdash A \fw B$, but using it
in a forward direction is impractical, because $B$ must be arbitrarily
assumed before knowing whether the rule is applicable.

More generally, in sequent formulations of syntactic calculi the goal
sequent for showing the grammaticality of a string $w_i$ has the form
\[
W_1 \cdots W_n \vdash S
\]
where $W_i$ gives the grammatical category of $w_i$ and $S$ is the
category of a sentence.  Proof search proceeds by matching current
sequents to the consequents of rules and trying to prove the
corresponding antecedents, or by recognizing a sequent as an axiom
instance $A\vdash A$. The corresponding natural deduction proof would
start from the assumptions $W_1, \ldots, W_n$ and try to prove $S$,
which is just the proof format that we have used here. However,
sequent rules like (\ref{right-slash}) above correspond to the
introduction of an additional assumption (not one of the $W_i$) at
some point in the proof and its later discharge, as in the
natural-deduction detachment rule for propositional logic. But such
undirected introduction of assumptions just in case they may yield
consequences that will be needed later is computationally very
costly.\footnote{There is more than a passing similarity between this
problem and the problem of pure bottom-up parsing with grammars with
gaps. In fact, a natural logical formulation of gaps is as assumptions
discharged by the {\it wh}-phrase they stand for
\cite{Pareschi+Miller:ICLP90,Hodas:gaps}.}
Systems that make full use of the sequent formulation therefore seem to
require top-down proof search.  It is of course possible to encode
top-down search in a bottom-up system by using more complex encodings
of search state, as is done in Earley's algorithm or in the magic
sets/magic templates compilation method for deductive databases
\cite{Bancilhon+Ramakrishnan:amateur,Ramakrishan:templates}.
Pentus \shortcite{Pentus93}, for instance, presents a compilation of Lambek
calculus to a CFG, which can then be processed by any of the standard
methods.  However, it is not clear yet that such techniques can be
applied effectively to grammatical sequent calculi so that they can be
implemented by the method described here.

\section{Control and Implementation}
\label{sec:control}

The specification of inference rules, as carried out in the previous
two sections, only partially characterizes a parsing algorithm, in
that it provides for what items are to be computed, but not in what
order.  This further control information is provided by choosing a
deduction procedure to operate over the inference rules.  If the
deduction procedure is complete, it actually makes little difference
in what order the items are enumerated, with one crucial exception: We
do not want to enumerate an item more than once.  To prevent this
possibility, it is standard to maintain a cache of lemmas, adding to
the cache only those items that have not been seen so far.  The cache
plays the same role as the {\it chart} in chart-parsing algorithms
\cite{kay-chart-parsing}, the {\it well-formed substring table} in CYK
parsing \cite{ka65,y67}, and the {\it state sets} in Earley's
algorithm \cite{e70}.  In this section, we develop a forward-chaining
deduction procedure that achieves this elimination of redundancy by
keeping a chart.

Items should be added to the chart as they are proved.  However, each
new item may itself generate new consequences.  The issue as to when
to compute the consequences of a new item is quite subtle.  A standard
solution is to keep a separate {\it agenda} of items that have been
proved but whose consequences have not been computed.  When an item is
removed from the agenda and added to the chart, its consequences are
computed and themselves added to the agenda for later consideration.

Thus, the general form of an agenda-driven, chart-based deduction
procedure is as follows:\label{sec:procedure}

\begin{enumerate}

\item Initialize the chart to the empty set of items and the agenda to
the axioms of the deduction system.

\item Repeat the following steps until the agenda is exhausted:

\begin{enumerate}

\item Select an item from the agenda, called the {\em trigger item},
and remove it.

\item Add the trigger item to the chart, if necessary.

\item If the trigger item was added to the chart, generate all items
that are new immediate consequences of the trigger item together with
all items in the chart, and add these generated items to the agenda.

\end{enumerate}

\item If a goal item is in the chart, the goal is proved (and the
string recognized); otherwise it is not.

\end{enumerate}

There are several issues that must be determined in making this
general procedure concrete, which we describe under the general topics
of {\em eliminating redundancy} and {\em providing efficient
access}. At this point, however, we will show that, under reasonable
assumptions, the general procedure is sound and complete.

In the arguments that follow, we will assume that items are always
ground and thus derivations are as defined in Section \ref{sec:pasd}. A
proof for the more general case, in which items denote sets of
possible grammaticality judgments, would require more intricate
definitions for items and inference rules, without changing the
essence of the argument.

\paragraph{Soundness}
We need to show that if the above procedure places item $I$ in the
chart when the agenda has been initialized in step~(1) with items
$A_1,\ldots,A_k$, then $A_1,\ldots,A_k\vdash I$. Since any item in the
chart must have been in the agenda, and been placed in the chart by
step~(2b), it is sufficient to show that $A_1,\ldots,A_k\vdash I$ for
any $I$ in the agenda. We show this by induction on the {\em stage}
$\sharp(I)$ of $I$, the number of the iteration of step~(2) at which
$I$ has been added to the agenda, or 0 if $I$ has been placed in the
agenda at step~(1).  Note that since several items may be added to the
agenda on any given iteration, many items may have the same stage
number.

If $\sharp(I)=0$, $I$ must be an axiom, and thus the trivial
derivation consisting of $I$ alone is a derivation of $I$ from
$A_1,\ldots,A_k$.

Assume that $A_1,\ldots,A_k\vdash J$ for $\sharp(J)<n$ and that
$\sharp(I)=n$. Then $I$ must have been added to the agenda by
step~(2c), and thus there are items $J_1,\ldots,J_m$ in the chart and
a rule instance such that
\[ \oneover{J_1 \quad \cdots \quad J_m}
           {I} \quad \mbox{$\langle${side conditions on
$J_1,\ldots,J_m,I$}$\rangle$}
\]
where the side conditions are satisfied. Since $J_1,\ldots,J_m$ are in
the chart, they must have been added to the agenda at the latest at
the beginning of iteration $n$ of step~(2), that is, $\sharp(J_i)<n$.
By the induction hypothesis, each $J_i$ must have a derivation
$\Delta_i$ from $A_1,\ldots,A_k$. But then, by definition of
derivation,
the concatenation of the derivations
$\Delta_1,\ldots,\Delta_m$ followed by $I$ is a derivation of $I$ from
$A_1,\ldots,A_k$.

\paragraph{Completeness}
We want to show that if $A_1,\ldots,A_k\vdash I$, then $I$ is in the chart
at step~(3). Actually, we can prove something stronger, namely that
$I$ is eventually added to the chart, if we assume some form of {\em
fairness} for the agenda. Then we will have covered cases in which
the full iteration of step~(2) does not terminate but step~(3) can
be interleaved with step~(2) to recognize the goal as soon as it
is generated. The form of fairness we will assume is that if
$\sharp(I)<\sharp(J)$ then item $I$ is removed from the agenda by step~(2a)
before item $J$. The agenda mechanism described in Section
\ref{sec:impl} below satisfies the fairness assumption.

We show completeness by induction on the length of any derivation
$D_1,\ldots,D_n$ of $I$ from $A_1,\ldots,A_k$.  (Thus we show
implicitly that the procedure generates every derivation, although in
general it may share steps among derivations.)

For $n=1$, $I=D_1=A_i$ for some $i$. It will thus be placed in the
agenda at step~(1), that is $\sharp(I)=0$. Thus by the fairness
assumption $I$ will be removed from the agenda in at most $k$
iterations of step~(2). When it is, it is either added to the chart as
required, or the chart already contains the same item.  (See
discussion of the ``if necessary'' proviso of step~(2b) in Section
\ref{sec:redundancy} below.)

Assume now that the result holds for derivations of length less than
$n$. Consider a derivation $D_1,\ldots,D_n=I$. Either $I$ is an axiom,
in which case we have just shown it will have been placed in the chart
by iteration $k$, or, by definition of derivation, there are
$i_1,\ldots,i_m<n$ such that there is a rule instance
\begin{equation}
\oneover{D_{i_1} \quad \cdots \quad D_{i_m}}
           {I} \quad \mbox{$\langle${side conditions on
		$D_{i_1},\ldots D_{i_m},I$}$\rangle$}
  \label{eqn:rule-app}
\end{equation}
with side conditions satisfied. By definition of derivation, each
prefix $D_1,\ldots,D_{i_j}$ of $D_1,\ldots,D_n$ is a derivation of
$D_{i_j}$ from $A_1,\ldots,A_k$. Then each $D_{i_j}$ is in the chart,
by the induction hypothesis. Therefore, for each $D_{i_j}$ there must
have been an identical item $I_j$ in the agenda that was added to the
chart at step~(2b). Let $I_p$ be the item in question that was the
last to be added to the chart. Immediately after that addition, all of
the $I_j$ (that is, all of the $D_{i_j}$) are in the chart, and
$I_p=D_{i_p}$ is the trigger item for rule application
(\ref{eqn:rule-app}). Thus $I$ is placed in the agenda. Since
step~(2c) can only add a finite number of items to the agenda, by the
fairness assumption item $I$ will eventually be considered at steps
(2a) and (2b), and added to the chart if not already there.

\subsection{Eliminating Redundancy}
\label{sec:redundancy}
\paragraph{Redundancy in the chart.}
The deduction procedure requires the ability to generate new
consequences of the trigger item and the items in the chart.  The key
word in this requirement is ``new''.  Indeed, the entire point of a
chart-based system is to allow caching of proved lemmas so that
previously proved (old) lemmas are not further pursued.  It is
therefore crucial that no item be added to the chart that already
exists in the chart, and it is for this reason that step~(2b) above
specifies addition to the chart only ``if necessary''.

\paragraph{Definition of ``redundant item''.}
The point of the chart is to serve as a cache of previously proved
items, so that an item already proved is not pursued.  What does it
mean for an item to be redundant, that is, occurring already in the
agenda or chart?  In the case of ground items, the appropriate notion
of occurrence in the chart is the existence of an identical chart
item.  If items can be non-ground (for instance, when parsing relative
to definite-clause grammars rather than context-free grammars) a more
subtle notion of occurrence in the chart is necessary.  As mentioned
above, a non-ground item stands for all of its ground instances, so
that a non-ground item occurs in the chart if all its ground instances
are covered by chart items, that is, if it is a specialization of some
chart item.  (This test suffices because of the {\em strong
compactness} of sets of terms defined by equations: if the instances
of a term $A$ are a subset of the union of the instances of $B$ and
$C$, then the instances of $A$ must be a subset of the instances of
either $B$ or $C$ \cite{Lassez+al:revisited}.) Thus, the appropriate
test is whether an item in the chart subsumes the item to be
added.\footnote{This subsumption check can be implemented in several
ways in Prolog.  \codeincluded{The code in the appendix presents two
of the options.}{The code made available with this paper presents two
of the options.}}

\paragraph{Redundancy in the agenda.} We pointed out that redundancy
checking in the chart is necessary.  The issue of redundancy in the
agenda is, however, a distinct one.  Should an item be added to the
agenda that already exists there?

Finding the rule that matches a trigger item, triggering the
generation of new immediate consequences, and checking that
consequences are new are expensive operations to perform. The
existence of duplicate items in the agenda therefore generates a
spurious overhead of computation especially in pathological cases
where exponentially many duplicate items can be created in the agenda,
each one creating an avalanche of spurious overhead.

For these reasons, it is also important to check for redundancy in the
agenda, that is, the notion of ``new immediate consequences'' in
step~(2c) should be interpreted as consequent items that do not
already occur in the chart {\em or agenda}.  If redundancy checking
occurs at the point items are about to be added to the agenda, it is
not required when they are about to be added to the chart; the ``if
necessary'' condition in step~(2b) will in this case be vacuous, since
always true.

\paragraph{Triggering the generation of new immediate consequences.}
With regard to step~(2c), in which we generate ``all items that are new
immediate consequences of the trigger item together with all other
items in the chart'', we would like, if at all possible, to refrain
from generating redundant items, rather than generating, checking for,
and disposing of the redundant ones.  Clearly, any item that is an
immediate consequence of the other chart items only (that is, without
the trigger item) is not a new consequence of the full chart.  (It
would have been generated when the last of the antecedents was itself
added to the chart.)  Thus, the inference rules generating new
consequences must have at least one of their antecedent items being
the trigger item, and the search for new immediate consequences can be
limited to just those in which at least one of the antecedents is the
trigger item.  The search can therefore be carried out by looking at
all antecedent items of all inference rules that match the trigger
item, and for each, checking that the other antecedent items are in
the chart.  If so, the consequent of that rule is generated as a
potential new immediate consequence of the trigger items plus other
chart items.  (Of course, it must be checked for prior existence in
the agenda and chart as outlined above.)

\subsection{Providing Efficient Access}

Items should be stored in the agenda and chart in such a way that they
can be efficiently accessed.  Stored items are accessed at two points:
when checking a new item for redundancy and when checking a
(non-trigger) antecedent item for existence in the chart.  For
efficient access, it is desirable to be able to directly index into
the stored items appropriately, but appropriate indexing may be
different for the two access paths.  We discuss the two types of
indexing separately, and then turn to the issue of variable renaming.

\paragraph{Indexing for redundancy checking.}
Consider, for instance, the Earley deduction system.  All items that
potentially subsume an item $[ i, A \ra \alpha \bullet \beta, j ]$
have a whole set of attributes in common with the item, for instance,
the indices $i$ and $j$, the production from which the item was
constructed, and the position of the dot (i.e., the length of
$\alpha$).  Any or all of these might be appropriate for indexing into
the set of stored items.

\paragraph{Indexing for antecedent lookup.} The information available
for indexing when looking items up as potential matches for
antecedents can be quite different.  In looking up items that match
the second antecedent of the completion rule $[ k, B \ra \gamma
\bullet {}, j]$, as triggered by an item of the form $[ i, A \ra
\alpha \bullet B \beta, k]$, the index $k$ will be known, but $j$ will
not be.  Similarly, information about $B$ will be available from the
trigger item, but no information about $\gamma$.  Thus, an appropriate
index for the second antecedent of the completion rule might include
its first index $k$ and the main functor of the left-hand-side $B$.
For the first antecedent item, a similar argument calls for indexing
by its second index $k$ and the main functor of the nonterminal $B$
following the dot.  The two cases can be distinguished by the sequence
after the dot: empty in the former case, non-empty in the latter.

\paragraph{Variable renaming.}
A final consideration in access is the renaming of variables.  As
non-ground items stored in the chart or agenda are matched against
inference rules, they become further instantiated.  This instantiation
should not affect the items as they are stored and used in proving
other consequences, so that care must be taken to ensure that
variables in agenda and chart items are renamed consistently before
they are used.  Prolog provides various techniques for achieving this
renaming implicitly.

\subsection{Prolog Implementation of Deductive Parsing}
\label{sec:impl}
In light of the considerations presented above, we turn now to our
method of implementing an agenda-based deduction engine in Prolog.  We
take advantage of certain features that have become standard in Prolog
implementations, such as clause indexing.  The code described below is
consistent with Quintus Prolog.

\subsubsection{Implementation of Agenda and Chart}

Since redundancy checking is to be done in both agenda and chart, we
need the entire set of items in both agenda and chart to be stored
together.  For efficient access, we store them in the Prolog database
under the predicate \term{stored/2}.  The agenda and chart are
therefore comprised of a series of unit clauses, e.g.,
\begin{center}
\begin{tabular}{ll}
  {\tt stored(1, item(...))}.		& $\longleftarrow$
						{\it beginning of chart}\\
 {\tt stored(2, item(...))}.			\\
 {\tt stored(3, item(...))}.			\\
 $\cdots$						\\
 {\tt stored($i-1$, item(...))}.	& $\longleftarrow$
						{\it end of chart}\\
 {\tt stored($i$, item(...))}.		& $\longleftarrow$
						{\it head of agenda}\\
 {\tt stored($i+1$, item(...))}.		\\
 $\cdots$						\\
 {\tt stored($k-1$, item(...))}.		\\
 {\tt stored($k$, item(...))}.		& $\longleftarrow$
						{\it tail of agenda}\\
\end{tabular}
\end{center}
The first argument of \term{stored/2} is a unique identifying index
that corresponds to the position of the item in the storage sequence
of chart and agenda items.  (This information is redundantly provided
by the clause ordering as well, for reasons that will become clear
shortly.)  The index therefore allows (through Quintus's indexing of
the clauses for a predicate by their first head argument) direct
access to any stored item.

Since items are added to the sequence at the end, all items in the
chart precede all items in the agenda.  The agenda items can therefore
be characterized by two indices, corresponding to the first ({\em
head}) and last ({\em tail}) items in the agenda.  A data structure
packaging these two ``pointers'' therefore serves as the proxy for the
agenda in the code.  An item is moved from the agenda to the chart
merely by incrementing the head pointer.  Items are added to the agenda
by storing the corresponding item in the database and incrementing the
tail pointer.

To provide efficient access to the stored items, auxiliary indexing
tables can be maintained.  Each such indexing table, is implemented as
a set of unit clauses that map access keys to the indexes of items
that match them.  In the present implementation, a single indexing
table (under the predicate \term{key_index/2}) is maintained that is
used for accessing items both for redundancy checking and for
antecedent lookup.  (This is possible because only the item attributes
available in both types of access are made use of in the keys, leading
to less than optimal indexing for redundancy checking, but use of
multiple indexing tables leads to much more database manipulation,
which is quite costly.)

In looking up items for redundancy checking, all stored items should be
considered, but for antecedent lookup, only chart items are pertinent.
The distinction between agenda and chart items is, under this
implementation, implicit.  The chart items are those whose index is
less than the head index of the agenda.  This test must be made
whenever chart items are looked up.  However, since clauses are stored
sequentially by index, as soon as an item is found that fails the test
(that is, is in the agenda) the search for other chart items can be
cut off.

\subsubsection{Implementation of the Deduction Engine}

Given the design decisions described above, the general agenda-driven,
chart-based deduction procedure presented in
Section~\ref{sec:procedure} can be implemented in Prolog as
follows:\codeincluded{\footnote{The code presented here diverges
slightly from that presented in the appendix for reasons of
exposition.}}{}

\beginprolog
parse(Value) :-
   {\em % (1) Initialize the chart and agenda}
    init_chart,
    init_agenda(Agenda),
   {\em % (2) Remove items from the agenda and process}
   {\em %     until the agenda is empty}
    exhaust(Agenda),
   {\em % (3) Try to find a goal item in the chart}
    goal_item_in_chart(Goal).
\endprolog

\noindent To exhaust the agenda, trigger items are repeatedly
processed until the agenda is empty:

\beginprolog
exhaust(Empty) :-
   {\em % (2) If the agenda is empty, we're done}
    is_empty_agenda(Empty).

exhaust(Agenda0) :-
   {\em % (2a) Otherwise get the next item index from the agenda}
    pop_agenda(Agenda0, Index, Agenda1),
   {\em % (2b) Add it to the chart}
    add_item_to_chart(Index),
   {\em % (2c) Add its consequences to the agenda}
    add_consequences_to_agenda(Index, Agenda1, Agenda),
   {\em % (2) Continue processing the agenda until empty}
    exhaust(Agenda).
\endprolog

For each item, all consequences are generated and added to the agenda:

\beginprolog
add_consequences_to_agenda(Index, Agenda0, Agenda) :-
    findall(Consequence,
            consequence(Index, Consequence),
            Consequences),
    add_items_to_agenda(Consequences, Agenda0, Agenda).
\endprolog

\noindent The predicate \term{add_items_to_agenda/3} adds the items
under appropriate indices as stored items and updates the head and
tail indices in \term{Agenda0} to form the new agenda \term{Agenda}.

A trigger item has a consequence if it matches an antecedent of some
rule, perhaps with some other antecedent items and side conditions,
and the other antecedent items have been previously proved (thus in
the chart) and the side conditions hold:

\beginprolog
consequence(Index, Consequent) :-
    index_to_item(Index, Trigger),
    matching_rule(Trigger,
                  RuleName, Others, Consequent, SideConds),
    items_in_chart(Others, Index),
    hold(SideConds).
\endprolog

\noindent Note that the indices of items, rather than the items
themselves are stored in the agenda, so that the index of the trigger
item must first be mapped to the actual item (with
\term{index_to_item/2}) before matching it
against a rule antecedent.  The \term{items_in_chart/2} predicate
needs to know both the items to look for (\term{Others}) and the index
of the current item (\term{Index}) as the latter distinguishes the
items in the chart (before this index) from those in the agenda (after
this index).

We assume that the inference rules are stored as unit clauses under
the predicate \term{inference(RuleName, Antecedents, Consequent,
SideConds)} where \term{RuleName} is some mnemonic name for the rule
(such as \term{predict} or \term{scan}), \term{Antecedents} is a list
of the antecedent items of the rule, \term{Consequent} is the single
consequent item, and \term{Sideconds} is a list of encoded Prolog
literals to execute as side conditions.  To match a trigger item
against an antecedent of an inference rule, then, we merely select a
rule encoded in this manner, and split up the antecedents into one
that matches the trigger and the remaining unmatched antecedents (to
be checked for in the chart).

\beginprolog
matching_rule(Trigger,
              RuleName, Others, Consequent, SideConds) :-
    inference(RuleName, Antecedents, Consequent, SideConds),
    split(Trigger, Antecedents, Others).
\endprolog

\subsubsection{Implementation of Other Aspects}

A full implementation of the deduction-parsing system --- complete
with encodings of several deduction systems and sample grammars --- is
\codeincluded{provided in the appendix}{available from
the first author}.  The \codeincluded{code in the
appendix}{distributed code} covers the following aspects of the
implementation that are not elsewhere described.

\begin{enumerate}

\item Input and encoding of the string to be parsed\codeincluded{
(Section~\ref{code:input})}{}.

\item Implementation of the deduction engine driver including
generation of consequences\codeincluded{
(Section~\ref{code:driver})}{}.

\item Encoding of the storage of items
\codeincluded{(Section~\ref{code:items}) }{}including the
implementation of the chart
\codeincluded{(Section~\ref{code:chart}) }{}and agenda\codeincluded{
(Section~\ref{code:agenda})}{}.

\item Encoding of deduction systems\codeincluded{
(Section~\ref{code:inference})}{}.

\item Implementation of subsumption checking\codeincluded{
(Section~\ref{code:utils})}{}.

\end{enumerate}

\noindent All Prolog code \codeincluded{is given in the Edinburgh
notation, and}{distributed} has been tested under the Quintus Prolog
system.

\subsection{Alternative Implementations}

This implementation of agenda and chart provides a compromise in terms
of efficiency, simplicity, and generality.  Other possibilities will
occur to the reader that may have advantages under certain conditions.
Some of the alternatives are described in this section.

\paragraph{Separate agenda and chart in database.}
Storage of the agenda and the chart under separate predicates in the
Prolog database allows for marginally more efficient lookup of chart
items; an extraneous arithmetic comparison of indices is eliminated.
However, this method requires an extra retraction and assertion when
moving an index from agenda to chart, and makes redundancy checking
more complex in that two separate searches must be engaged in.

\paragraph{Passing agenda as argument.}
Rather than storing the agenda in the database, the list of agenda
items might be passed as an argument.  (The implementation of queues
in Prolog is straightforward, and would be the natural structure to
use for the agenda argument.)  This method again has the marginal
advantage in antecedent lookup, but it becomes almost impossible to
perform efficient redundancy checking relative to items in the agenda.

\paragraph{Efficient bottom-up interpretation.}
The algorithm just presented can be thought of as a pure bottom-up
evaluator for inference rules given as definite clauses, where the
head of the clause is the consequent of the rule and the body is the
antecedent. However, given appropriate inference rules, the bottom-up
procedure will simulate non-bottom-up parsing strategies, such as the
top-down and Earley strategies described in Section
\ref{sec:cf}. Researchers in deductive databases have extensively
investigated variants of that idea: how to take advantage of the
tabulation of results in the pure bottom-up procedure while keeping
track of goal-directed constraints on possible answers. As part of
these investigations, efficient bottom-up evaluators for logic
programs have been designed, for instance CORAL
\cite{Ramakrishnan+al:coral}. Clearly, one could use such a system
directly as a deduction parser.

\paragraph{Construction of derivations.} \label{sec:imp-offline}
The direct use of the inference rules for building derivations, as
presented in Section~\ref{sec:offline}, is computationally
inefficient, since it eliminates structure-sharing in the chart.  All
ways of deriving the same string will yield distinct items, so that
sharing of computation of subderivations is no longer possible.

A preferable method is to compute the derivations offline by
traversing the chart after parsing is finished.  The deduction engine
can be easily modified to do so, using a technique reminiscent of that
used in the Core Language Engine \cite{cle}.  First, we make use of
two versions of each inference rule, an online version such as the
Earley system given in Figure~\ref{tab:summary}, with no computation
of derivations, and an offline version like the one in
Figure~\ref{fig:earley-mod} that does generate derivation information.
We will presume that these two versions are stored, respectively,
under the predicates \term{inference/4} (as before) and
\term{inference_offline/4}, with the names of the rules specifying the
correspondence between related rules.  Similarly, the online
\term{initial_item/1} specification should have a corresponding
\term{initial_item_offline/1} version.

The deduction engine parses a string using the online version of the
rules, but also stores, along with the chart, information about the
ways in which each chart item was constructed, with unit clauses of
the form
\beginprolog
stored_history(Consequent, Rule, Antecedents).\eqpunc{,}
\endprolog
\noindent which specifies that the item whose index is given by
\term{Consequent} was generated by the inference rule whose name is
\term{Rule} from the antecedent items given in the sequence
\term{Antecedents}.  (If an item is generated as an initial item, its
history would mark the fact by a unit clause using the constant
\term{initial} for the \term{Rule} argument.)

When parsing has completed, a separate process is applied to each
goal item, which traverses these stored histories using the second
(offline) version of the inference rules rather than the first,
building derivation information in the process.  The following Prolog
code serves the purpose.  It defines
\term{offline_item(Index, Item)}, a predicate that computes the
offline item \term{Item} (presumably including derivation information)
corresponding to the online item with index given by \term{Index},
using the second version of the inference rules, by following the
derivations stored in the chart history.

\beginprolog
offline_item(Index, Item) :-
    stored_history(Index, initial, _NoAntecedents),
    initial_item_offline(Item).
offline_item(Index, Item) :-
    stored_history(Index, Rule, Antecedents),
    offline_items(Antecedents, AntecedentItems)
    inference_offline(Rule, AntecedentItems, Item, SideConds),
    hold(SideConds).

offline_items([], []).
offline_items([Index|Indexes], [Item|Items]) :-
    offline_item(Index, Item),
    offline_items(Indexes, Items).
\endprolog

The offline version of the inference rules need not merely compute a
derivation.  It might perform some other computation dependent on
derivation, such as semantic interpretation.  Abstractly, this
technique allows for staging the parsing into two phases, the second
comprising a more fine-grained version of the first.  Any staged
processing of this sort can be implemented using this technique.

\paragraph{Finer control of execution order}

For certain applications, it may be necessary to obtain even finer
control over the order in which the antecedent items and side
conditions are checked when an inference rule is triggered.  Given
that the predicates \term{items_in_chart/2} and \term{holds/1} perform
a simple left-to-right checking of the items and side conditions, the
implementation of \term{matching_rule/5} above leads to the remaining
antecedent items and side conditions being checked in left-to-right
order as they appear in the encoded inference rules, and the side
conditions being checked after the antecedent items.  However, it may
be preferable to check antecedents and side conditions interleaved,
and in different orders depending on which antecedent triggered the
rule.

For instance, the side condition $A=A'$ in the second inference rule
of Section~\ref{sec:augmented} must be handled before checking for the
nontrigger antecedent of that rule, in order to minimize
nondeterminism.  If the first antecedent is the trigger, we want to
check the side conditions and then look for the second antecedent, and
correspondingly for triggering the second antecedent.  The
implementation above disallows this possibility, as side conditions
are always handled after the antecedent items.  Merely swapping the
order of handling side conditions and antecedent items, although
perhaps sufficient for this example, does not provide a general
solution to this problem.

Various alternatives are possible to implement a finer level of
control.  We present an especially brutish solution here, although
more elegant solutions are possible.  Rather than encoding an
inference rule as a single unit clause, we encode it with one clause
per trigger element under the predicate
\beginprolog
inference(RuleName, Antecedents, Consequent)
\endprolog
\noindent where \term{Rulename} and \term{Consequent}
are as before, but \term{Antecedents} is now a list of all the
antecedent items and side conditions of the rule, with the trigger
item first.  (To distinguish antecedent items from side conditions, a
disambiguating prefix operator can be used, e.g., \term{@item(...)}
versus \term{?side_condition(...)}.)  Matching an item against a rule
then proceeds by looking for the item as the first element of this
antecedent list.

\beginprolog
matching_rule(Trigger, RuleName, Others, Consequent) :-
    inference(RuleName, [Trigger|Others], Consequent),
\endprolog

\noindent The \term{consequence/2} predicate is modified to use this
new \term{matching_rule/4} predicate, and to check that all of the
antecedent items and side conditions hold.

\beginprolog
consequence(Index, Consequent) :-
    index_to_item(Index, Trigger),
    matching_rule(Trigger, RuleName, Others, Consequent),
    hold(Others, Index).
\endprolog

The antecedent items and side conditions are then checked in the order
in which they occur in the encoding of the inference rule.

\beginprolog
hold([], _Index).
hold([Antecedent|Antecedents], Index) :-
    holds(Antecedent, Index),
    hold(Antecedents, Index).

holds(@Item,      Index) :- item_in_chart(Item, Index).
holds(?SideCond, _Index) :- call(SideCond).
\endprolog

\section{Conclusion}

The view of parsing as deduction presented in this paper, which
generalizes that of previous work in the area, makes possible a simple
method of describing a variety of parsing algorithms --- top-down,
bottom-up, and mixed --- in a way that highlights the relationships
among them and abstracts away from incidental differences of control.
The method generalizes easily to parsers for augmented phrase
structure formalisms, such as definite-clause grammars and other logic
grammar formalisms.  Although the deduction systems do not specify
detailed control structure, the control information needed to turn
them into full-fledged parsers is uniform, and can therefore be given
by a single deduction engine that performs sound and complete
bottom-up interpretation of the rules of inference.  The implemented
deduction engine that we described has proved useful for rapid
prototyping of parsing algorithms for a variety of formalisms
including variants of tree-adjoining grammars, categorial grammars,
and lexicalized context-free grammars.

\section*{Acknowledgements}

This material is based in part upon work supported by the National
Science Foundation under Grant No. IRI-9350192 to SMS.  The authors
would like to thank the anonymous reviewers for their helpful comments
on an earlier draft.

\bibliographystyle{fullname}
%\bibliography{infer}

\appendix

\newpage

\newlength{\codewidth}
\setlength{\codewidth}{\textwidth}
\addtolength{\codewidth}{-.75in}

\newcommand{\codelist}[1]{
\begin{center}
\rule{\codewidth}{.2pt}\\
$\langle${\sc Listing of file }{\tt #1.pl}$\rangle$
\end{center}
{\small
\prologlisting{Code/#1.pl}}
\begin{center}
$\langle${\sc End listing of file }{\tt #1.pl}$\rangle$\\
\rule{\codewidth}{.2pt}
\end{center}
}

\section{Full Code for the Deductive Parsing Engine}

\label{sec:full-code}

\codelist{infer}

\subsection{Reading and Encoding of Input} \label{code:input}

It is standard in the logic grammar literature to use a list encoding
of strings and string positions.  A string is taken to be a list of
words, with string positions encoded as the suffix of that list
beginning at the index in question.  Thus, position 3 of the string
encoded \term{[terry, writes, a, program, that, halts]} would be
encoded by the list \term{[a, program, that, halts]}.  Items, which
include one or more such string positions, can become quite large, and
testing for identity of string positions cumbersome, especially as
these items are to be stored directly in the Prolog database.  For
this reason, we use a more direct encoding of string positions, and a
commensurately more complicated encoding of the underlying strings.
String positions will be taken to be integers.  A string will then be
encoded as a series of unit clauses (using the predicate \term{word/2}
specifying which words occur before which string positions.  For
instance, the string ``Terry writes a program that halts'' would be
specified with the unit clauses

\beginprolog
word(1, terry).
word(2, writes).
word(3, a).
word(4, program).
word(5, that).
word(6, halts).
\endprolog

The end of the string is no longer explicitly represented in this
encoding, so that a predicate \term{sentencelength/1} will be used to
specify this information.

\beginprolog
sentencelength(6).
\endprolog

A predicate to read in an input string can perform the conversion to
this encoded form automatically, asserting the appropriate unit
clauses as the string is read in.

\codelist{input}

\subsection{Deduction Engine Driver} \label{code:driver}

The main driver operates as per the discussion in
Section~\ref{sec:control}.

\codelist{driver}

\subsection{Stored Items Comprising Chart and Agenda} \label{code:items}

\codelist{items}

\subsubsection{Chart Items} \label{code:chart}

\codelist{chart}

\subsubsection{Agenda Items} \label{code:agenda}

The agenda items are just a contiguous subsequence of the stored
items.  The specification of which items are in the agenda (and
therefore which are implicitly in the chart) is provided by the head
and tail indices of the agenda subsequence.  This queue specification
of the agenda, packed into a term under the functor \term{queue/2}, is
passed around as an argument by the deduction engine.  A term
\term{queue(Head, Tail)} represents a queue of agenda items where
\term{Head} is the index of the first element in the queue and
\term{Tail} is the index of the next element to be put in the queue
(one more than the current last element).

Notice the asymmetry between enqueueing and dequeueing: enqueueing
(\term{add_item_to_agenda/3}) takes explicit items, dequeueing
(\term{pop_agenda/3}) produces indices that may then be mapped to
items by \term{index_to_item/2}.  This is somewhat inelegant, but
balances the need for abstraction in the generic algorithm with the
efficiency of the main \term{all_solutions} there, which need not
store the item whose consequences are being sought.

This implementation is adequate because the items in the agenda always
form a contiguous set of stored items, so their indices are
sequential.

\codelist{agenda}

\subsection{Encoding of Deductive Parsing Systems}
\label{code:inference}

We present the Prolog encodings of several of the deduction systems
discussed above including all of the context-free systems from
Section~\ref{sec:pasd} and the CCG system described in
Section~\ref{sec:ccg}.

The deduction systems for context-free-based (definite clause)
grammars all assume the same encoding of a grammar as a series of unit
clauses of the following forms:

\begin{description}

\item[Grammar rules:]  Grammar rules are encoded as clauses of the
form \term{LHS ---> RHS} where \term{LHS} (a nonterminal) and
\term{RHS} (a list of nonterminals and preterminals) are respectively
the left- and right-hand side of a rule.

\item[Lexicon:]  The lexicon is encoded as a relation between
preterminals and terminals by unit clauses of the form \term{lex(Term,
Preterm)}, where \term{Preterm} is a preterminal (nonterminal
dominating a terminal) that covers the \term{Terminal}.

\item[Start symbol:]  The start symbol of the grammar is encoded by a
unit clause of the form \term{startsymbol(Start)}, where \term{Start}
is a start nonterminal for the grammar; there may be several such
nonterminals.

\end{description}

Nonterminals and terminal symbols are encoded as arbitrary terms and
constants.  The distinction between nonterminals and terminals is
implicit in whether or not the terms exist on the left-hand side of
some rule.

\codelist{inference}

\subsubsection{The Top-Down System}

\codelist{inf-top-down}

\subsubsection{The Bottom-Up System}

\codelist{inf-bottom-up}

\subsubsection{The Earley's Algorithm System}

\codelist{inf-earley}

\subsubsection{The Combinatory Categorial Grammar System}

The CCG parser in this section assumes an encoding of the CCG
grammar/lexicon as unit clauses of the form \term{lex(Word,
Category)}, where \term{Word} is a word in the lexicon and
\term{Category} is a CCG category for that word.  Categories are
encoded as terms using the infix functors \term{+} for forward slash
and \term{-} for backward slash.

The start category is encoded as for context-free grammars above.

\codelist{inf-ccg}

\subsection{Sample Grammars}

\codelist{grammars}

\subsubsection{A Sample Definite-Clause Grammar}

The following is the definite-clause grammar of Figure~\ref{fig:toy}
encoded as per Section~\ref{code:inference}.

\codelist{gram-dcg}

\subsubsection{A Sample Combinatory Categorial Grammar}

The following is the combinatory categorial grammar of
Figure~\ref{fig:ccgtoy} encoded appropriately for the CCG deduction
system.

\codelist{gram-ccg}

\subsection{Utilities}
\label{code:utils}

\codelist{utilities}

\subsection{Monitoring and Debugging}

\codelist{monitor}

\end{document}